\documentclass{article}

\usepackage{microtype}
\usepackage{graphicx}
\usepackage{subcaption}
\usepackage{booktabs} 

\usepackage{hyperref}

\usepackage[preprint]{icml2026}

\usepackage{amsmath}
\usepackage{amssymb}
\usepackage{mathtools}
\usepackage{amsthm}

\usepackage[capitalize,noabbrev]{cleveref}

\theoremstyle{plain}

\theoremstyle{definition}

\theoremstyle{remark}

\usepackage[textsize=tiny]{todonotes}

\usepackage{subcaption}
\usepackage{algorithmic}
\usepackage{placeins}
\usepackage{wrapfig}
\usepackage[table]{xcolor}
\usepackage{bm}
\usepackage{multirow}

\usepackage{algorithm}

\DeclareMathOperator*{\argmax}{arg\,max}

\newcommand{\STAB}[1]{\begin{tabular}{@{}c@{}}#1\end{tabular}}

\icmltitlerunning{Adapting Time Series Foundation Models through Data Mixtures}

\begin{document}

\twocolumn[
  \icmltitle{Adapting Time Series Foundation Models through Data Mixtures}



  \icmlsetsymbol{equal}{*}

  \begin{icmlauthorlist}
    \icmlauthor{Thomas~L. Lee}{inf}
    \icmlauthor{Edoardo M.~Ponti}{inf}
    \icmlauthor{Amos Storkey}{inf}
  \end{icmlauthorlist}

  \icmlaffiliation{inf}{School of Informatics, University of Edinburgh}

  \icmlcorrespondingauthor{Thomas~L. Lee}{t.l.lee-1@sms.ed.ac.uk}

  \icmlkeywords{Machine Learning, ICML}

  \vskip 0.2in
]



\printAffiliationsAndNotice{}  

\begin{abstract}
Time series foundation models (TSFMs) have become increasingly popular for zero-shot forecasting. However, for a new time series domain not fully covered by the pretraining set, performance can suffer.
Therefore, when a practitioner cares about a new domain and has access to a set of related datasets, the question arises: \textit{how best to fine-tune a TSFM to improve zero-shot forecasting?} 
A typical approach to this type of problem is to fine-tune a LoRA module on all datasets or separately on each dataset. Tuning a separate module on each dataset allows for the specialisation of the TSFM to different types of data distribution, by selecting differing combinations of per-dataset modules for different time series contexts. However, we find that, using per-dataset modules might not be optimal, since a time series dataset can contain data from several types of distributions, i.e. \textit{sub-domains}. This can be due to the distribution shifting or having differing distributions for different dimensions of the time series. 
Hence, we propose \textit{MixFT} which re-divides the data using Bayesian mixtures into sets that best represent the sub-domains present in the data, and fine-tunes separately on each of these sets. This re-division of the data ensures that each set is more homogeneous, leading to fine-tuned modules focused on specific sub-domains. 
Our experiments show that MixFT performs better than per-dataset methods and when fine-tuning a single module on all the data. This suggests that by re-partitioning the data to represent sub-domains we can better specialise TSFMs to improve zero-shot forecasting. 
\end{abstract}

\section{Introduction}

\begin{figure*}[!t]
        \centering
        \includegraphics[width=\textwidth,trim={0 50cm 35cm 0},clip]{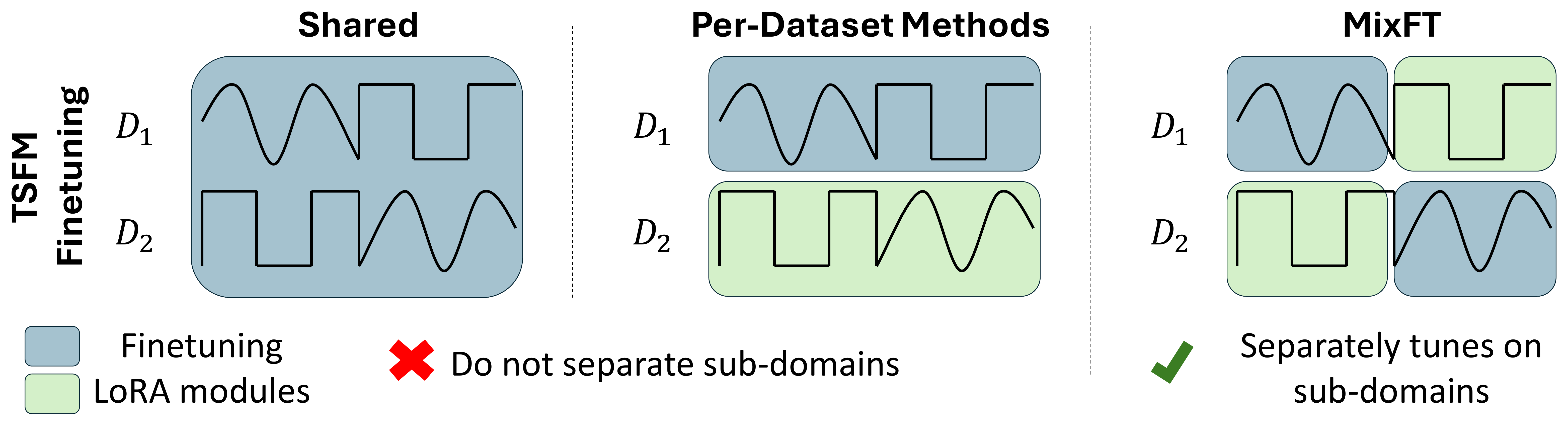}
        \caption{\textbf{MixFT identifies and trains separate LoRA modules on different sub-domains given in the fine-tuning datasets.} This is unlike previous approaches which either separately fine-tune LoRA modules using dataset boundaries (per-dataset methods) or trained a single LoRA module on all the data (Shared) \citep{ostapenko2024towards}. By training separate LoRA modules on sub-domains, MixFT aims for the training data for each LoRA to be more homogeneous, leading to more specialized and consistent LoRA modules. This also should lead to better identifiability of what LoRA modules to use when zero-shot forecasting. }
        \label{fig:1}
\end{figure*}
Time series foundation models (TSFMs) have become increasingly popular for forecasting \citep{miller2024survey, ye2024survey, liang2024foundation}. A key reason for this is their zero-shot forecasting ability \citep{Ansari2024Chronos, Woo2024moirai, darlow2024dam}. This allows for the production of forecasts where there is little historical data, expertise or compute to fit a model on the particular time series of interest \citep{mercier2021evaluating}. However, while there might not be much historical data for a particular time series, often a practitioner will have access to other related time series. Hence, a question arises of how to best adapt the TSFM with these other related time series to improve zero-shot performance on the time series of interest. This question is of particular importance when the time series of interest is from a domain outside of the pretraining set of the TSFM. This is because, it has been observed that zero-shot performance can suffer greatly for these time series \citep{tonerperformance, diao2024forecasting, aksu2024gift}, requiring the need for TSFM adaptation to obtain useful forecasts. 

There are several potential ways to adapt a TSFM given a set of time series datasets. In this paper, we focus on parameter-efficient fine-tuning (PEFT) methods, due to their strong performance in LLMs \citep{han2024parameter} and recently for TSFMs \citep{gupta2024low, gupta2024beyond}. The most straightforward way is to fine-tune a LoRA adapter  \citep{hu2021lora} on all of the available fine-tuning datasets, which is called \textit{shared} fine-tuning \citep{prabhu2023computationally}. By looking at work done in fine-tuning LLMs, we find that the other common way to fine-tune across multiple datasets is to train separate LoRA modules on each dataset or on sets of datasets \citep{ostapenko2024towards}. These methods can be grouped under the term \textit{per-dataset methods} \citep{ostapenko2024towards, chronopoulou2023adaptersoup}. Per-dataset methods have the advantage that each LoRA module can specialise on modelling the data distributions occurring in the datasets it is trained on, which has been shown to lead to better performance \citep{ponti2023combining}. 

However, a time series often has distribution shifts and different dimensions of a multivariate time series dataset might belong to significantly different distributions \citep{zhang2024addressing, kim2025battling}, i.e. sub-domains. Hence, we identify that there can be a better way to divide the data than using dataset labels, by having the data of each sub-domain be in a different division. More formally, we observe that we can separate the data distribution across time series datasets into a set of sub-distributions. Here, a sub-distribution represents a particular abstract type of time series data, for example, different seasonal patterns or levels of spikiness. We call each sub-distribution a \textit{sub-domain}. The goal of this paper is to explore this approach of identifying and separately training LoRA modules on the sub-domains of the data instead of per-dataset.

We propose a method \textit{MixFT} which identifies and fine-tunes using better divisions of the fine-tuning data than per-dataset ones. MixFT leverages Bayesian mixture models \citep{murphy2022probabilistic} to learn and identify sub-domains and hence split the data based on sub-domains rather than dataset boundaries. MixFT then trains a LoRA module \textit{per-sub-domain}. This ensures that each LoRA module is specialised to a sub-domain, simplifying learning as the data each LoRA module is trained on is more homogenous. By using Bayesian mixture models, we also have the ability when zero-shot forecasting to identify the time series context's sub-domain and hence what LoRA module to use for forecasting. By doing this, we ensure that the data used for fine-tuning is \textit{in (sub)-distribution} to that being forecasted. We find in our experiments that MixFT performs well, outperforming both per-dataset fine-tuning approaches and when fine-tuning on all the data together. This suggests that by fitting LoRA modules on sub-domains rather than datasets, they can specialise better as the training data is more consistent, leading to more accurate forecasts. 

This work has the following three main contributions:
\begin{enumerate}
    \item Identifying that dataset divisions can often be not optimal when fine-tuning specialized LoRA modules for zero-shot forecasting.  

    \item Proposal of MixFT, which leverages Bayesian mixture models to identify better data divisions for fine-tuning based on the sub-domains of the data. Using Bayesian mixtures also allows for, when zero-shot forecasting, the identification of a time series contexts sub-domain. This enables the selection of the LoRA module which has been trained on data of the same sub-distribution as the context.

    \item The study of the performance of fine-tuning methods on TSFMs, showing that MixFT performs well and that some per-dataset methods perform poorly compared to not fine-tuning at all. 
\end{enumerate}

\section{Related Work}
In this work we look at time series foundation models (TSFMs). Current TSFMs are usually large transformer-based architectures trained on a large amount of time series data \citep{rasul2023lag,Das2024TimesFM,liang2024foundation}. For example, Chronos Bolt \citep{Ansari2024Chronos} is a TSFM based on the T5 architecture and was trained on around 100 billion time series data instances. Another popular line of TSFMs are the Moirai models \citep{Woo2024moirai} which use masked encoder architectures. While TSFMs are becoming increasingly popular, compared to other types of foundation models like LLMs there has currently been little work on fine-tuning methodologies \citep{miller2024survey, leelightweight}. For example there has been little work on exploring PEFT for TSFMs \citep{gupta2024low, gupta2024beyond}. Additionally, most work looking at fine-tuning TSFMs focus on improving performance on the current time series \citep{Ekambaram2024Tiny}. This is in contrast to this work which instead focuses on fine-tuning on a collection of datasets to improve zero-shot forecasting on related time series.   

While there has been relatively little work on how to fine-tune TSFMs to improve zero-shot performance, in the LLM literature it has been well studied \citep{han2024parameter}. For example, there has been significant work looking at using PEFT for zero-shot prediction \citep{ostapenko2024towards}. A general approach is to train a LoRA module on each of the fine-tuning datasets and then form a linear combination of them to perform zero-shot prediction \citep{chronopoulou2023adaptersoup}. Additionally, there have also been RAG-based, prompt-based and in-context-based methods proposed for the adaptation of foundation models to improve zero-shot performance \citep{gao2023retrieval, liu2023pre, dong2022survey}. However, these methods are often specifically tailored to LLMs and require especially trained TSFMs for their use \citep{yang2025timerag, ning2025ts, das2024context}. This is one of the reasons we adopt LoRA modules as the basis of MixFT, as we aim for it to be agnostic to the choice of TSFM. 
Further afield, there has also been work on selecting subsets of data to improve generalisation by improving the interference between the gradients of different data instances \citep{maharana2024adapt, swayamdipta2020dataset, gadre2023datacomp}. 


\section{Preliminaries: Time Series Forecasting}
\label{sec:prelims}
To describe MixFT and our experiments we first need to lay the groundwork of time series forecasting and fine-tuning TSFMs. A time series consists of a sequence of time indexed vectors $\bm{v}_{0}, \ldots, \bm{v}_{t}, \ldots, \bm{v}_{T}$ where $\bm{v}_t \in \mathbb{R}^c$ is the value of the time series at time step $t$. In general the value of a time series can be multivariate, i.e. $c>1$, and we look at such time series in our experiments. However, for simplicity and without loss of generality, below and in Section~\ref{sec:methodology}, we only present the univariate case, i.e. $c=1$. 

When fine-tuning and zero-shot forecasting we consider $L$-length context windows of a time series,
\begin{align}
    \bm{x}_{i, j}=[\bm{v}_{i, t}]_{t=j}^{j+L},
\end{align}
where $i$ identifies the time series. This windowing is standard practice in the literature \citep{darlow2023foldformer, lim2021time, nietime2022}. From each of the $M$ time series available for fine-tuning we construct a windowed time series dataset $D_i = [\bm{x}_{i, j} ]_{j=1}^{N_i}$, for $i \in [1, \ldots, M]$. Here $N_i$ denotes the number of context windows for this time series, $N_i=T_i-L$, where $T_i$ is the length of the time series. These windowed time series are then used by a fine-tuning method to adapt the TSFM. 

\textit{Zero-shot forecasting} consists of giving the TSFM is a context $\bm{x}$ from a \textit{new} time series, not seen during fine-tuning \citep{aksu2024gift}. Its task is to generate a forecast $\hat{\bm{y}}$ of length $H$ which predicts the future values of the time series after $\bm{x}$ has been observed. In evaluation, forecasts are compared to the true values of the time series for the time steps of that forecast, denoted as $\bm{y}$. The reason windows are used is that it converts the time series data into the common format used in machine learning (ML), of covariates $\bm{x}$ and targets $\bm{y}$, such that standard ML approaches can be leveraged \citep{lim2021time, toneranalysis}.    

\subsection{Notation}
We list the notation used throughout the paper below. 
\begin{itemize}
    \item $L$: Context length (number of time steps in the input).
    \item $H$: Forecast horizon (number of future time steps to forecast).
    \item $M$: Number of fine-tuning time series datasets.
    \item $N_i$: Number of context windows in time series $i$.
    \item $K$: Number of components used by MixFT
    \item $\bm{x}_{i,j}$: The $j$th context window from the $i$th time series, $\bm{x}_{i,j} \in \mathbb{R}^{L}$.
    \item $\bm{z}_{i,j}$: Embedded context window, $\bm{z}_{i,j} = \text{TSFM}_{embed}(\bm{x}_{i,j})$
    \item $\bm{y}, \bm{\hat{y}}$: Target and forecast respectively, $\bm{y}, \bm{\hat{y}} \in \mathbb{R}^{H}$ for some context window $\bm{x}$.
\end{itemize}

\section{MixFT: Fine-Tuning with Dataset Mixtures}
\label{sec:methodology}
\begin{figure*}[!t]
        \centering
        \includegraphics[width=\textwidth,trim={0 27cm 0 0},clip]{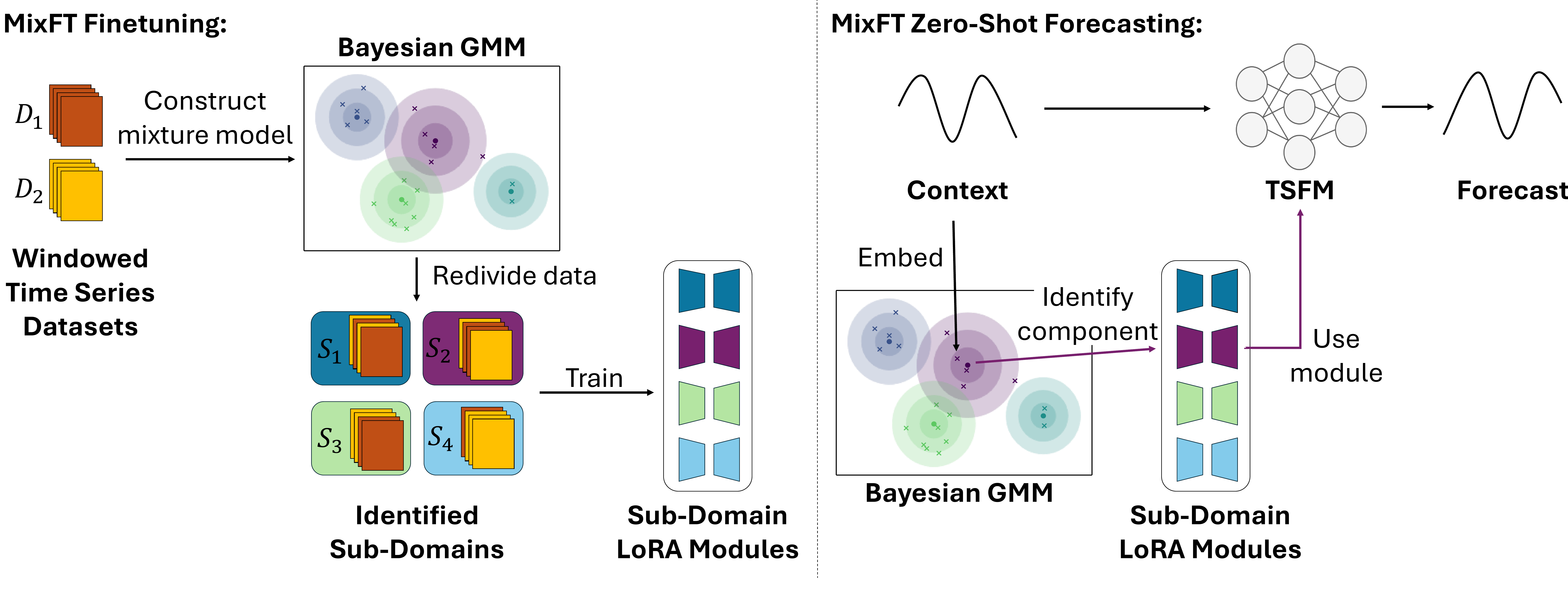}
        \caption{\textbf{Overview of MixFT.} On the left side of the figure, we display how MixFT fine-tunes LoRA modules for zero-shot forecasting. First, it identifies the sub-domains in the fine-tuning datasets. Then it redivides the data per-sub-domain. Last, it trains a separate LoRA module on the data of each sub-domain. This ensures that the trained LoRA modules specialise in forecasting sub-domains not datasets, unlike previous work \citep{ostapenko2024towards}. On the right side of the figure, we show how MixFT constructs a forecast. This is done by identifying what sub-domain a context belongs to and then using that sub-domains LoRA module for forecasting. This ensures the TSFM is exploiting the knowledge of that sub-domain given by the fine-tuning data.}
        \label{fig:Overview}
        \vspace{1mm}
\end{figure*}
In this work, we propose \textit{MixFT}, a TSFM fine-tuning method for zero-shot forecasting. MixFT is different from previous work in that it redivides the datasets provided as fine-tuning data into sets that better align with the underlying sub-domains of the data. To achieve this, MixFT fits a Bayesian Gaussian mixture model (GMM) to the fine-tuning data to identify the sub-domains. Then, MixFT splits the data into sets belonging to each of the mixture components and tunes a separate LoRA module on each set. This constitutes the fine-tuning stage of MixFT. To perform zero-shot forecasting, given a time series context, MixFT first identifies the mostly likely sub-domain/mixture component it belongs to. Then, MixFT uses the particular LoRA module trained for that subdomain and constructs a forecast. Both the fine-tuning stage and the zero-shot forecasting of MixFT are described in detail below and diagrammatically in Figure~\ref{fig:Overview}---algorithmic descriptions are also presented in Appendix~\ref{appen:algo}.       
  
\subsection{Fine-Tuning}
MixFT fine-tuning consists of two stages: \textbf{a)} fitting a Bayesian GMM and \textbf{b)} fitting LoRA modules on the data partitions given by the Bayesian GMM. To perform the fine-tuning MixFT is given a set of windowed time series of a context length $L$ (see Section~\ref{sec:prelims} for details). More formally, MixFT is given $D_i = [\bm{x}_{i, j} ]_{j=1}^{N_i}$ for $i \in [1, \ldots, M]$. The Bayesian GMM is defined in the embedding space of the TSFM, to alleviate the need for a more complex model. Hence, MixFT first embeds the data using the TSFM ($\text{TSFM}_{embed}$), resulting in the embedded datasets $D^Z_i = [\bm{z}_{i, j} = \text{TSFM}_{embed}(\bm{x}_{i,j}) ]_{j=1}^{N_i}$. Now, we can define the Bayesian GMM MixFT uses, which has $K$ mixture components,  
\begin{align}
    \bm{\mu}_k, \bm{\Sigma}_k &\sim \text{NIDW}(\bm{m}, \kappa, \nu, \bm{W}) \;  \; \text{ for $k \in [1, \ldots,K]$}, \\
    \bm{\pi} &\sim \text{Dir}(\bm{\alpha}), \\
    c_{i, j} &\sim \text{Cat}(\bm{\pi}), \\
    z_{i,j} | c_{i, j} &\sim \mathcal{N}(\bm{\mu}_{c_{i, j}}, \bm{\Sigma}_{c_{i, j}}),
\end{align}
where $\bm{m}, \kappa, \nu, \bm{W}$ and $\bm{\alpha}$ are hyperparameters and NIDW denotes the normal-inverse-diagonal-Wishart distribution (the inverse-diagonal-Wishart distribution is the set of per-dimension independent inverse chi-squared distributions \citep{murphy2022probabilistic}). By using a NIDW prior, the covariance $\bm{\Sigma}_k$ for each mixture component is diagonal, keeping the parameters count low and reducing any chance of overfitting (see Appendix~\ref{appen:mixmodel}). 

We fit the Bayesian GMM using mean-field variational inference \citep[Section~10.2]{BishopPRML2006}, where we assume the latent variables $c_{i,j}$ are independent of $\bm{m}$, $\bm{W}$ and $\bm{\pi}$ for our variational distribution. By performing variational inference, MixFT has access to the posterior variational distributions,
\begin{align}
    q(\bm{\mu}_k, \bm{\Sigma}_k) &= \text{NIDW}(\bm{m}_k^{\text{VAR}}, \kappa_k^{\text{VAR}}, \nu_k^{\text{VAR}}, \bm{W}_k^{\text{VAR}}), \\
    q(\bm{\pi}) &= \text{Dir}(\bm{\alpha}^{\text{VAR}}),
\end{align}
defined by the estimates $\bm{m}_k^{\text{VAR}}, \kappa_k^{\text{VAR}}, \nu_k^{\text{VAR}}, \bm{W}_k^{\text{VAR}}$ and $\bm{\alpha}^{\text{VAR}}$. We use variational inference due its increased stability over the maximum likelihood estimation (i.e. K-means) \citep{murphy2022probabilistic}, ensuring the found mixture is less dependant on its initialisation. Additionally, we explored the use of more complex probabilistic models, in particular embedded topic models \citep{dieng2020topic} but found they had an increased chance to overfit compared to Bayesian GMMs (see Appendix~\ref{appen:mixmodel}). 

Given access to the posterior variational distribution, we can now redivide the data into sets that align with the latent sub-domains of the data. To do this, we give a sub-domain label for each datapoint using the posterior predictive distribution,
\begin{align} \label{eq:postPred}
    \hat{c}_{i,j} &= \argmax_{c \in [1,\ldots,K]} [p(c | \bm{z}_{i,j}, \phi_c^{\text{VAR}}, \bm{\alpha}^{\text{VAR}})].
\end{align}
Where $\phi_c^{\text{VAR}} = \{\bm{m}_c^{\text{VAR}}, \kappa_c^{\text{VAR}}, \nu_c^{\text{VAR}}, \bm{W}_c^{\text{VAR}}\}$. Then we split the data into sub-domain sets,
\begin{align}
    \mathcal{S}_k = \{\bm{x}_{i,j} | \hat{c}_{i,j} = k\},
\end{align}
for $k$ in range 1 to $K$. 

The sets $\mathcal{S}_k$ are then used for the second stage of MixFT fine-tuning: training of the sub-domain LoRA modules. To do this we simply train a separate LoRA module on each set $\mathcal{S}_k$. This results in MixFT having a library of $K$ LoRA modules, each specialised to a specific sub-domain found in the data. 

MixFT fits LoRA modules this way for several reasons. First, it ensures that each LoRA component needs only to model the characteristics of its given sub-domain, simplifying its task. Second, it reduces the possibility of destructive interference which can happen when jointly modelling different competing sub-domains/tasks, as shown by \citet{liu2021conflict}. Last, by repartitioning the data, the sub-domain expert LoRA modules MixFT creates reduces the \textit{generalisability gap} in zero-shot forecasting \citep{mehta2025generalization, he2024right}. This is because, for a given sub-domain, there is less of a distribution shift between the data the sub-domains LoRA was trained on and that sub-domains data used in zero-shot forecasting, then when using per-dataset LoRAs. The reason for this is that the per-dataset LoRA have been trained on a variety of data from differing sub-domains and so any per-dataset LoRA must be further away in distribution to a new time series context sampled from a sub-domain than MixFT sub-domains LoRA module. The reduction in the generalisability gap leads to better zero-shot forecasting performance, as evidenced by our experiments (see Section~\ref{sec:exps}). 

\subsection{Zero-Shot Forecasting} 
When zero-shot forecasting, we are given a time series context $\bm{x}$ from a related but different time series from the fine-tuning data. The way that it is related is by assuming the time series roughly contains the same latent sub-domains as our fine-tuning data, else there would be no reason to fine-tune. Given this, MixFT's first task is to identify which sub-domain $\bm{x}$ belongs to. By using a Bayesian GMM, we have an immediate way to do this, by using the most likely sub-domain given by the fitted posterior predictive distribution,
\begin{align}
    \hat{c} &= \argmax_{c \in [1,\ldots,K]} [p(c | \bm{z},  \phi_c^{\text{VAR}}, \bm{\alpha}^{\text{VAR}})] 
\end{align}
where $\bm{z}=\text{TSFM}_{embed}(\bm{x})$ and $\phi_c^{\text{VAR}} = \{\bm{m}_c^{\text{VAR}}, \kappa_c^{\text{VAR}}, \nu_c^{\text{VAR}}, \bm{W}_c^{\text{VAR}}\}$ are MixFT's variational estimates. Now that we have an estimate of what sub-domain $\bm{x}$ is from, to give its forecast we use the $\hat{c}$th LoRA module which models the $\hat{c}$th sub-domain. MixFT uses the LoRA module with the TSFM and input $\bm{x}$, producing a forecast $\hat{y}$. 

By forecasting this way we aim for $\bm{x}$ to be more \textit{in distribution} of the data fine-tuned on by the selected LoRA module, being of the same sub-domain. This in turn should mean the LoRA module can use the learnt characteristics of that sub-domain to aid the TSFM in giving a more accurate forecast. The reason we use a \textit{hard} sub-domain assignment, taking the argmax over the posterior predictive, for a given context, is due to the combination of two reasons: \textbf{a)} the forecasts given when using LoRA modules of unlikely sub-domains can be inaccurate due to the out of distribution (OOD) nature of the context for that sub-domain (see Appendix~\ref{appen:ent}) and \textbf{b)} in our experiments we find that MixFT is usually certain what sub-domain a given context is from (see Appendix~\ref{appen:topic_forecasts}). Hence, by only using the forecast given by using the LoRA of the most likely sub-domain, which is usually much more probable than the rest in our experiments, MixFT is not effected by inaccurate OOD forecasts from using LoRAs of unlikely sub-domains. Additionally, previous methods often use a post-hoc way to select LoRA modules or mixtures thereof. This can lead to problems where the fine-tuning method and selection method are misaligned. However, MixFT's use of a probabilistic model overcomes this problem as it directly uses how the LoRA modules were fine-tuned to decide how they are selected for zero-shot forecasting. 

\section{Experiments}
\label{sec:exps}
\subsection{Experimental Setup}
\begin{table*}[t!]
\setlength\tabcolsep{2.2pt}
\setlength{\extrarowheight}{0.7pt}
\centering
\caption{\textbf{MASE of fine-tuning methods.} The table shows the MASE performance of fine-tuning methods on the Chronos Bolt and Moirai-1.1-R TSFMs, when zero-shot forecasting standard time series datasets. A lower MASE is better and we also provide standard errors over training runs. Importantly, the time series used for fine-tuning are related but are not the same as the time series evaluated on. We bold and underline the best and second best performing method, respectively, for each dataset and TSFM. Our results show that MixFT, in terms of average ranking and number of bolded/underlined results, performs the best. }
\label{table:results_main}
\begin{tabular}{@{}cc@{\hskip 1mm}|@{\hskip 1mm}ccccccc@{}}
\toprule
\addlinespace
\multicolumn{1}{c}{} & \multicolumn{1}{c}{} & \multicolumn{7}{c}{Methods} \\
\cmidrule(l){3-9}
\multirow{2}{*}{} & \multirow{2}{*}{Datasets} &  \multirow{2}{*}{Base} & \multirow{2}{*}{Shared}  &  \multirow{2}{*}{\begin{tabular}{c} $\mu$ \\ Datasets \\ \end{tabular}} &  \multirow{2}{*}{\begin{tabular}{c} Arrow \\ Datasets \\ \end{tabular}} &  \multirow{2}{*}{Poly} & \multirow{2}{*}{MBC} &  \multirow{2}{*}{\begin{tabular}{c} MixFT \\ (ours) \\ \end{tabular}} \\ 
\multirow{2}{*}{} & \\
\midrule
\multirow{12}{*}{\STAB{\rotatebox[origin=c]{90}{\textbf{Chronos Bolt}}}} 
& CloudD1 & $1.511$ & $1.553_{\pm{0.024}}$ & $1.580_{\pm{0.016}}$ & $\mathbf{1.352_{\pm{0.006}}}$ & $\underline{1.463_{\pm{0.012}}}$ & $1.810_{\pm{0.266}}$ & $1.485_{\pm{0.008}}$ \\
& CloudD2 & $1.572$ & $1.342_{\pm{0.004}}$ & $1.614_{\pm{0.029}}$ & $\underline{1.341_{\pm{0.005}}}$ & $\mathbf{1.320_{\pm{0.004}}}$ & $1.649_{\pm{0.211}}$ & $1.343_{\pm{0.004}}$ \\
\addlinespace
& BizITObs-L2C & $3.116$ & $\underline{3.101_{\pm{0.011}}}$ & $3.624_{\pm{0.084}}$ & $3.680_{\pm{0.028}}$ & $3.124_{\pm{0.001}}$ & $3.170_{\pm{0.009}}$ & $\mathbf{3.083_{\pm{0.006}}}$ \\
& BizITObs-App & $3.494$ & $\underline{1.060_{\pm{0.020}}}$ & $3.691_{\pm{0.117}}$ & $5.002_{\pm{0.197}}$ & $1.173_{\pm{0.006}}$ & $1.419_{\pm{0.346}}$ & $\mathbf{0.989_{\pm{0.024}}}$ \\
& US-Births & $1.014$ & $\underline{0.967_{\pm{0.007}}}$ & $1.175_{\pm{0.003}}$ & $1.212_{\pm{0.003}}$ & $0.979_{\pm{0.010}}$ & $1.041_{\pm{0.005}}$ & $\mathbf{0.942_{\pm{0.002}}}$ \\
\addlinespace
& M4-Daily & $\mathbf{7.175}$ & $7.254_{\pm{0.016}}$ & $8.415_{\pm{0.088}}$ & $8.376_{\pm{0.037}}$ & $7.249_{\pm{0.004}}$ & $7.530_{\pm{0.121}}$ & $\underline{7.238_{\pm{0.005}}}$ \\
& M4-Monthly & $7.670$ & $\mathbf{7.580_{\pm{0.042}}}$ & $9.032_{\pm{0.506}}$ & $10.658_{\pm{0.141}}$ & $7.735_{\pm{0.013}}$  & $8.173_{\pm{0.129}}$ & $\underline{7.638_{\pm{0.014}}}$\\
& M4-Quarterly & $\underline{8.178}$ & $8.269_{\pm{0.037}}$ & $10.916_{\pm{0.187}}$ & $9.089_{\pm{0.052}}$ & $8.253_{\pm{0.037}}$ & $8.275_{\pm{0.123}}$ & $\mathbf{8.135_{\pm{0.010}}}$ \\
\addlinespace
& ETTh2 & $\mathbf{1.383}$ & $1.388_{\pm{0.001}}$ & $1.414_{\pm{0.004}}$ & $1.431_{\pm{0.003}}$ & $\underline{1.386_{\pm{0.001}}}$ & $1.390_{\pm{0.002}}$ & $1.387_{\pm{0.001}}$ \\
& ETTm2 & $\mathbf{0.723}$ & $0.731_{\pm{0.003}}$ & $0.787_{\pm{0.002}}$ & $0.785_{\pm{0.003}}$ & $0.731_{\pm{0.001}}$ & $0.740_{\pm{0.003}}$ & $\underline{0.730_{\pm{0.001}}}$ \\
\addlinespace
\cmidrule(l){2-9}
\rowcolor[gray]{0.8} & \textbf{Avg. Rank ($\mathbf{\downarrow}$)} & $2.9$ & $3$ & $6.3$ & $5.6$ & $\underline{2.8}$ & $5.3$ & $\mathbf{2}$ \\

\midrule
\addlinespace
\addlinespace

\multirow{12}{*}{\STAB{\rotatebox[origin=c]{90}{\textbf{Moirai-1.1-R}}}} 
& CloudD1 & $2.164_{\pm{0.001}}$ & $1.814_{\pm{0.016}}$ & $1.923_{\pm{0.015}}$ & $\mathbf{1.596_{\pm{0.006}}}$ & $2.183_{\pm{0.094}}$ & $2.079_{\pm{0.051}}$ & $\underline{1.813_{\pm{0.011}}}$ \\
& CloudD2 & $2.522_{\pm{0.001}}$ & $1.832_{\pm{0.004}}$ & $1.844_{\pm{0.023}}$ & $\mathbf{1.506_{\pm{0.004}}}$ & $\underline{1.721_{\pm{0.032}}}$ & $2.052_{\pm{0.032}}$ & $1.774_{\pm{0.015}}$ \\
\addlinespace
& BizITObs-L2C & $3.262_{\pm{0.001}}$ & $\underline{3.166_{\pm{0.004}}}$ & $4.420_{\pm{0.060}}$ & $3.903_{\pm{0.028}}$ & $3.174_{\pm{0.010}}$ & $3.191_{\pm{0.013}}$ & $\mathbf{3.149_{\pm{0.001}}}$ \\
& BizITObs-App & $5.127_{\pm{0.001}}$ & $3.011_{\pm{0.084}}$ & $5.313_{\pm{0.186}}$ & $5.926_{\pm{0.171}}$ & $\underline{1.574_{\pm{0.068}}}$ & $\mathbf{1.400_{\pm{0.056}}}$ & $2.762_{\pm{0.004}}$ \\
& US-Births & $1.641_{\pm{0.001}}$ & $1.663_{\pm{0.014}}$ & $2.139_{\pm{0.026}}$ & $1.859_{\pm{0.027}}$ & $\underline{1.618_{\pm{0.004}}}$ & $\mathbf{1.597_{\pm{0.005}}}$ & $1.647_{\pm{0.009}}$ \\
\addlinespace
& M4-Daily & $\mathbf{7.740_{\pm{0.001}}}$ & $7.813_{\pm{0.007}}$ & $9.602_{\pm{0.114}}$ & $9.102_{\pm{0.061}}$ & $7.866_{\pm{0.016}}$ & $7.978_{\pm{0.084}}$ & $\underline{7.767_{\pm{0.003}}}$ \\
& M4-Monthly & $8.012_{\pm{0.001}}$ & $\underline{7.962_{\pm{0.013}}}$ & $10.645_{\pm{0.259}}$ & $9.988_{\pm{0.155}}$ & $8.173_{\pm{0.172}}$  & $8.077_{\pm{0.085}}$ & $\mathbf{7.876_{\pm{0.008}}}$ \\
& M4-Quarterly & $\mathbf{7.777_{\pm{0.002}}}$ & $\underline{7.975_{\pm{0.002}}}$ & $9.171_{\pm{0.068}}$ & $8.809_{\pm{0.035}}$ & $8.083_{\pm{0.039}}$ & $8.163_{\pm{0.033}}$ & $7.998_{\pm{0.018}}$ \\
\addlinespace
& ETTh2 & $1.498_{\pm{0.001}}$ & $\underline{1.475_{\pm{0.002}}}$ & $1.586_{\pm{0.005}}$ & $1.546_{\pm{0.004}}$ & $1.485_{\pm{0.002}}$ & $1.483_{\pm{0.003}}$ & $\mathbf{1.474_{\pm{0.001}}}$ \\
& ETTm2 & $0.948_{\pm{0.001}}$ & $\underline{0.848_{\pm{0.001}}}$ & $0.890_{\pm{0.002}}$ & $0.868_{\pm{0.001}}$ & $0.866_{\pm{0.008}}$ & $0.861_{\pm{0.005}}$ & $\mathbf{0.846_{\pm{0.001}}}$ \\
\addlinespace
\cmidrule(l){2-9}
\rowcolor[gray]{0.8} & \textbf{Avg. Rank ($\mathbf{\downarrow}$)} & $4.3$ & $\underline{2.9}$ & $6.3$ & $5.0$ & $3.7$ & $3.7$ & $\mathbf{2.1}$ \\
\bottomrule
\vspace{1mm}
\end{tabular}
\end{table*}

\textbf{Benchmark} \: To analyse the performance of MixFT and, more generally, whether redividing fine-tuning data into more homogeneous sets gives performance benefits, we perform an number of experiments. In our experiments we have used datasets from the Cloud and Gift-Eval benchmarks \citep{tonerperformance, aksu2024gift}. These benchmarks are known to be challenging to zero-shot forecast well by current TSFMs and contain many closely related datasets. 

We have split the datasets from each benchmark into two groups, fine-tuning data and evaluation data. Specifically, we use for fine-tuning data the datasets: CloudD3, CloudD4, BizITObs-Service, BitBrains Fast Storage (hourly), M4-Hourly and M4-Weekly \citep{tonerperformance, palaskar2024automixer, ShenStat2024, godahewa2monash}. For evaluation we use the datasets: CloudD1, CloudD2, BizITObs-L2C, BizITObs-App, US-Births, M4-Daily, M4-Monthly, M4-Quarterly, ETTh2 and ETTm2 \citep{tonerperformance, palaskar2024automixer, godahewa2monash, zhou2021informer}. We note here that we evaluate on all the data in each of the evaluation datasets. This is instead of using predefined test splits sometimes used in previous work \citep{aksu2024gift}. We have used a context length $L=520$ and a prediction length of $H=30$ in our experiments, as commonly used in previous work \citep{Ansari2024Chronos, Ekambaram2024Tiny}. The reason behind the structure of our setup is that it simulates the real-world use-case we target: fine-tuning on a range of related datasets to improve the TSFMs zero-shot performance on time series a practitioner cares about. 

The TSFMs we evaluate on are Chronos Bolt (small) and Moirai-1.1-R (small) \citep{Ansari2024Chronos, Woo2024moirai}. We have selected these TSFM as they have state-of-the-art performance and are not pretrained on any of the fine-tuning or evaluation data we have used \citep{aksu2024gift}. We present additional experimental details in Appendix~\ref{appen:ExpDetails}.  

\textbf{Fine-tuning methods} \: We compare MixFT to several other fine-tuning methods designed to improve zero-shot performance. We first compare to the most straightforward method \textit{Shared} fine-tuning which trains a single LoRA module on all of the available fine-tuning data \citep{ostapenko2024towards}. We also compare to several per-dataset methods: \textit{$\mu$-Datasets} \citep{chronopoulou2023adaptersoup, page2023multi}, which tunes a separate LoRA module on each dataset and then uses the average of the LoRA modules when zero-shot forecasting; \textit{Arrow-Dataset} \citep{ostapenko2024towards}, which uses the well-performing \textit{Arrow} zero-shot routing mechanism to weight per-dataset adapters for each evaluation time series context given; \textit{Poly} \citep{prabhu2023computationally}, which is an end-to-end approach, where in training, there is learnt importance weighting for each dataset for each LoRA module used; and \textit{MBC} \citep{ostapenko2024towards}, which aggregates datasets together using statistics from per-dataset LoRAs and then trains a LoRA module on each aggregate. We also compare to the baseline \textit{Base} which is when we do not fine-tune the TSFM and instead use its pretraining weights for zero-shot forecasting. For MixFT, MBC and Poly we use $K=2$ LoRA modules/data divisions in our experiments, unless stated otherwise. We present in Appendix~\ref{appen:K_selection}, a method for automatically selecting $K$, using average rank on a validation set, which validates our choice of $K=2$.

Each fine-tuning method looked at tunes a set of LoRA modules \citep{hu2021lora}, but ensuring the modules do not overfit is challenging \citep{gupta2024low, gupta2024beyond}.\footnote{But, less challenging than fine-tuning all the parameters of a TSFM \citep{leelightweight}} To mitigate this problem, we use two strategies \textbf{a)} MixUp data augmentations \citep{zhang2017MixUp, Ansari2024Chronos} and \textbf{b)} retraining with pretraining data alongside the new fine-tuning data, a method known as experience replay \citep{ostapenko2022continual, de2021continual}. By using both strategies we aim to not overfit to the specific fine-tuning data nor damage the retention of pretraining performance. Instead our goal is to focus the TSFM on the data distributions present in the fine-tuning data, so to improve zero-shot performance on related data.   

\textbf{Metrics} \: We use the \textit{mean absolute scaled error} (MASE) to measure the accuracy of forecasts \citep{hyndman2006another}. MASE is defined for a given forecast $\bm{\hat{y}}$, context $\bm{x}$ and target $\bm{y}$ as
\begin{align}
    \text{MASE}(\bm{\hat{y}}, \bm{y}, \bm{x}) = \frac{L-S}{H} \frac{\sum_{i=1}^{H} |\hat{y}_i - y_i|}{\sum_{i=1}^{L-S}|x_i-x_{i+S}|},
\end{align}
where $S$ represents the seasonality. MASE is a normalised mean absolute error (MAE), where the MAE of the navie seasonal forecaster on the context is used for normalisation. Normalising enables for better aggregation of the performance of forecasts for time series which vary in numerical scale and complexity. Because of this, MASE is commonly used in previous works on TSFMs \citep{Ansari2024Chronos} and is the reason why we use it. 

To give a summary of the performance of each fine-tuning method, we record the average rank in each of our results tables. The \textit{average rank} of a method is the average over its placing on each of the datasets. For example, an average rank of $2$ would mean on average the method was second best over the datasets. We use average rank as it treats each dataset equally and so is not effected by the different variances in performance over methods between datasets.   

\subsection{Results}
We presents the results of comparing different fine-tuning methods in Table~\ref{table:results_main}. We find that MixFT performs the best. In the table we present the zero-shot performance across the evaluation datasets, bolding and underlining the best and second best perform methods for each dataset and TSFM, respectively. We find, in terms of average rank and the number of best and second best performances, that MixFT performs better than any of the other fine-tuning methods. Additionally, we find that many fine-tuning methods struggle to perform better than not fine-tuning (Base). In fact, apart from MixFT, for Chronos Bolt, the better performing TSFM, each method is outperformed by Base by at least $0.01$ MASE for at least 3 datasets. MixFT is outperformed by Base by at least $0.01$ MASE only for M4-Daily. This is unlike the results in the LLM literature, where fine-tuning methods nearly always outperform Base \citep{ostapenko2024towards}, showing the difficulty of fine-tuning TSFMs for zero-shot performance.  

\begin{table}[t!]
\setlength{\extrarowheight}{0.6pt}
\centering
\caption{\textbf{Average MASE improvement over all datasets on Chronos Bolt for fine-tuning methods using arrow routing when compared to Base (i.e. not fine-tuning).} We present both the average improvement in terms of absolute values (Avg. Imp) and relative values (Avg. Rel. Imp.). Relative values are calculated as the MASE of Base divided by the MASE of the fine-tuning method in question. We find that when controlling for how the LoRA modules are selected at test time, by always using arrow routing, the LoRA modules learnt by MixFT perform the best. This suggests that it finds better divisions of the fine-tuning data than per-dataset divisions.}
\label{table:topics}
\begin{tabular}{@{}c@{\hskip 1mm}|@{\hskip 1mm}ccccc@{}}
\toprule
\addlinespace
\multirow{2}{*}{Methods} & \multirow{2}{*}{Avg. Imp.($\downarrow$)}  & \multirow{2}{*}{Avg. Rel. Imp.($\uparrow$)} \\ 
\\
\midrule
Arrow-Datasets & $0.71$  & $0.90$x\\
MBC & $\underline{-0.06}$ &  \underline{$1.11$x} \\
Arrow-MixFT & $\mathbf{-0.17}$ & $\mathbf{1.17}$\textbf{x} \\
\bottomrule
\end{tabular}
\vspace{-1mm}
\end{table}

In Table~\ref{table:topics} we analyse further the difference between fine-tuning on the data mixtures found by MixFT and fine-tuning on datasets or sets of datasets. To do this, we control for how the fine-tuned LoRA modules are selected at test time by exclusively using the Arrow selection method. We compare the LoRA modules found by MixFT (Arrow-MixFT) to when training a LoRA module per dataset (Arrow-Datasets) or over sets of datasets (MBC). By only using Arrow selection/routing, we ensure that the performance difference between the methods is not due to better selection at test time but due to better trained LoRA modules. Table~\ref{table:topics} presents the results of this experiment, showing the LoRA modules trained on MixFTs data mixtures perform better than per-dataset ones. This suggests that by redividing the fine-tuning data into sets that align more with the underlying sub-domains of the data, MixFT trains better LoRA modules. 


\subsection{What Mixtures does MixFT Learn?}

\begin{figure}[t]
    \centering
    \includegraphics[width=\columnwidth]{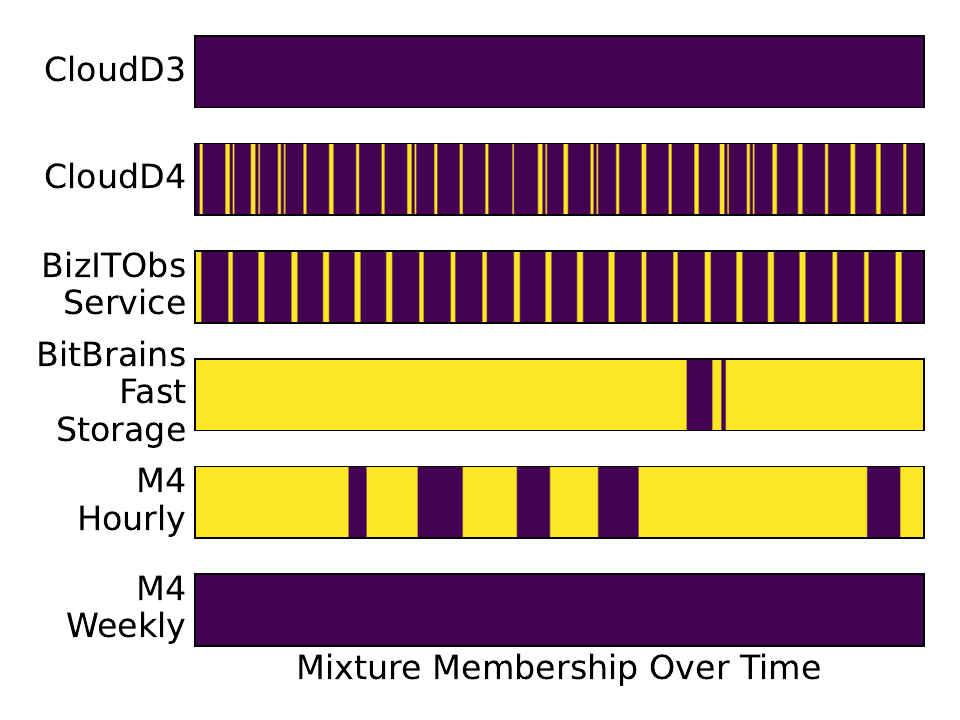}
    \caption{\textbf{Mixture membership of the fine-tuning datasets for MixFT.} The plot shows, for the first channel of each fine-tuning dataset, when the time series is learnt to be of the first mixture component (purple area) or the second component (yellow area). We find that some of the channels consist of just one mixture component (CloudD3 and M4-Weekly), mostly one component (BitBrains and M4-Hourly) or exhibit periodic patterns (CloudD4 and BizITObs-Service). The plots suggest that MixFTs Bayesian mixture model finds reasonable patterns in the data. Also, as both sub-domains/mixture components are found in a given dataset, MixFT's data divisions can not be found by per-dataset methods. }
    \label{fig:mixDist}
    \vspace{1mm}
\end{figure}

It is valuable to explore what dataset divisions/mixtures MixFT finds to better understand its behaviour and performance. To do this we plot in Figure~\ref{fig:mixDist} what mixture component MixFT assigns to the data of the first channel for each fine-tuning dataset across time. We find that both mixture components are used frequently and that for the CloudD3/4, BizITObs-Service and M4-weekly datasets the first (purple) component is preferred while for BitBrains Fast Storage and M4-Hourly the other component used more frequently. This shows different usages of the mixture components across datasets. Additionally, for CloudD4 and BizITObs-Service we observe a periodic pattern of mixture assignment. This suggests MixFT's mixtures have identified some seasonal patterns/sub-domains. Also, in Appendix~\ref{appen:ent} we look at how confident MixFT is in assigning mixture components. We find that for both the fine-tuning data and the evaluation data MixFT is highly confident at identifying what mixture component to use for a given context. Furthermore, in Appendix~\ref{Appen:contexts}, we look at time series contexts identified as belonging to the different sub-domains. We find that there are distinct characteristics belonging to each sub-domain. For instance, whether at the end of a context the time series is changing rapidly or not. 

It is important to note that, MixFT uses both mixture components for the same dataset (see Figure~\ref{fig:mixDist}), which is impossible to achieve with per-dataset methods. This showcases MixFT flexibility and ability to identify multiple sub-domains inside the same dataset. This is a key benefit of using MixFT instead of a per-dataset method, leading to better fine-tuned LoRA modules (see Table~\ref{table:topics}) and superior zero-shot forecasting.

\subsection{Ablations}
To understand the usefulness of different components of MixFT we perform several ablations. First, we look at MixFT's method to select LoRA modules when zero-shot forecasting. We compare to using $\mu$ and Arrow selections methods as well as using the probabilities given by the mixture model. The results are presented in Appendix~\ref{appen:selection} and show that MixFT's selection method performs the best. This is because, MixFT performs best or second best over all datasets, apart from CloudD1 where it is third best and achieves the lowest average rank. Second, we look at using different probabilistic models to redivide the data, in particular a GMM fitted by K-means \citep{murphy2022probabilistic} and an embedding topic model \citep{dieng2020topic}. The results and details of this experiment can be found in Appendix~\ref{appen:mixmodel}. We find that the use of Bayesian GMMs by MixFT outperforms the other two models looked at. Specifically, we find it performs the best or second best for all bar one of the datasets (M4-Daily). This suggests that MixFTs Bayesian GMM is well suited to finding sub-domains in time series data. Last, we analyse the number of mixture components used in Appendix~\ref{appen:K_selection} and \ref{appen:components}. We find general agreement between the performance of using MixFT with different number of components on validation data (see Appendix~\ref{appen:K_selection}) and on evaluation data (see Appendix~\ref{appen:components}), where using $K=2$ performs best for both. This suggests the method proposed for selecting $K$ finds a reasonable value.  

\section{Conclusions}
In this work we have proposed MixFT, a fine-tuning method which uses Bayesian mixtures to divide the fine-tuning data into its latent sub-domains. By doing so, it can train a LoRA module on each sub-domain, making it specialised to the characteristics of that sub-domain while not potentially confusing it with data from others. This allows for better zero-shot forecasting. This is because, MixFT can use the LoRA weights fine-tuned on a sub-domain to predict new data from that sub-domain. This reduces the generalisation gap between the fine-tuning and testing data compared to methods where each LoRA module is trained on multiple sub-domains, such as per-dataset methods. We demonstrate the benefits of MixFT in our experiments---we also list its limitations in Appendix~\ref{appen:limits}. We find that it has superior zero-shot forecasting performance to per-dataset methods or when training a single LoRA module on all the fine-tuning data. Therefore, we see MixFT as suggesting a potential direction for future TSFM fine-tuning methods, based on how to best to compartmentalise the available fine-tuning data, instead of seeing the dataset partitions as fixed. 




\section*{Impact Statement}


This paper presents work whose goal is to improve time series forecasting. There are many potential societal consequences of our work. However, these consequences are not particular to our work and instead are shared across machine learning as a whole. Therefore, we feel that none of these
consequences must be specifically highlighted here.


\bibliography{References}
\bibliographystyle{icml2026}

\newpage
\appendix
\onecolumn

\section{Limitations}
\label{appen:limits}
Throughout the main text we have aimed to have highlighted the design decisions and limitations of MixFT. In addition to this, we would like to present here the main limitations of MixFT. The first limitation is that we assume the sub-domains are shared between the fine-tuning and evaluation data. Hence, when this is not the case we would not expect MixFT to perform well. However, we note that in our experiments that the sharing of sub-domains seems to be the case, given MixFT's good performance. Furthermore, it is \textit{necessary} to assume some relationship between fine-tuning and evaluation data to make any claims about zero-shot generalisation \citep{gouk2022limitations}. Therefore, while we make explicit our assumption of how we link fine-tuning and evaluation data any similar method must also make some assumption, explicit or implicit. Last, the real world use case we are targeting is that of a practitioner selecting and using fine-tuning data to improve zero-shot forecasting for a particular type of time series. Hence, the practitioner should know why they think the fine-tuning data chosen relates to the time series they want to evaluate on and hence if the assumptions of MixFT are valid. 

The second limitation we would like to mention is the memory/parameter and computational overhead of using MixFT. In terms of the added memory, or parameters, needed by MixFT on top of the base model, MixFT needs to store one LoRA module per sub-domain and a negligible amount of memory for the Bayesian GMM. This is roughly the same as Poly and MBC, which are state-of-the-art methods compared to in our experiments. Additionally, it is less than Arrow-Dataset which uses a LoRA module for each dataset. On the other hand, the Shared and $\mu$-Datasets methods do not use any more parameters/memory than the base model when zero-shot forecasting. This is because, it is possible adsorb the single LoRA model these methods use into the base model. In terms of added compute cost, when fine-tuning, MixFT has to fit the Bayesian GMM which can take some time and hence is slower to train than shared fine-tuning. But, we found that MixFT roughly had the same training time as other methods like MBC. When forecasting, MixFT has to classify the sub-domain of the context, this adds computational overhead, requiring the base models feature representation of the context. Hence, MixFT requires more time to produce forecasts than using shared fine-tuning. However, the speed at which forecasts are produced by MixFT is less than the sampling frequency of datasets used in our experiments, reducing the impact of MixFT's computational overhead. 

\section{Additional Methodological Details}
\subsection{MixFTs Fine-Tuning and Zero-Shot Forecasting Algorithms}
\begin{algorithm}[h]
   \caption{\textbf{: MixFT Zero-Shot Forecasting}}
   \label{alg:forecast}
\begin{algorithmic}
   \STATE {\bfseries Input:} Time series context $\bm{x}$; per-sub-domain LoRA modules $[\text{LoRA}_k]_{k=1}^{K}$; Bayesian GMM $[\bm{m}_k^{\text{VAR}}, \kappa_k^{\text{VAR}}, \nu_k^{\text{VAR}}, \bm{W}_k^{\text{VAR}}]_{k=1}^{K}, \; \bm{\alpha}^{\text{VAR}}$; and TSFM 
   \STATE 
   \STATE Identify sub-domain of $\bm{x}$:
    \STATE  $\hat{c} = \argmax_{c \in [1,\ldots,K]} [p(c | \bm{z}  = \text{TSFM}_{embed}(\bm{x}), \bm{m}_k^{\text{VAR}}, \kappa_k^{\text{VAR}}, \nu_k^{\text{VAR}}, \bm{W}_k^{\text{VAR}}, \bm{\alpha}^{\text{VAR}})]$
   \STATE
   \STATE Compute and return TSFM forecast using LoRA module of the $\hat{c}$th sub-domain:
   \STATE $\bm{\hat{y}} = \text{Forecast}(\text{TSFM}, \; \text{LoRA}_c, \; \bm{x})$
   \STATE{\bfseries Return} $\bm{\hat{y}}$
\end{algorithmic}
\end{algorithm}

\label{appen:algo}
\begin{algorithm}[h]
   \caption{\textbf{: MixFT Fine-Tuning}}
   \label{alg:finetune}
\begin{algorithmic}
   \STATE {\bfseries Input:} Fine-tuning datasets $D_i = [\bm{x}_{i, j} ]_{j=1}^{N_i}$ for $i \in [1, \ldots, M]$; TSFM ($ \text{TSFM}_{embed}$); and number of mixture components $K$
   \STATE 
   \STATE Embed datasets:
   \FOR{$i \in [1, \ldots, M]$}
   \STATE $D^Z_i = [\bm{z}_{i, j} = \text{TSFM}_{embed}(\bm{x}_{i,j}) ]_{j=1}^{N_i}$
   \ENDFOR
   \STATE
   \STATE Fit Bayesian GMM:
   \STATE $[\bm{m}_k^{\text{VAR}}, \kappa_k^{\text{VAR}}, \nu_k^{\text{VAR}}, \bm{W}_k^{\text{VAR}}]_{k=1}^{K}, \; \bm{\alpha}^{\text{VAR}} = \text{BayesGMM}(\bigcup_{i=1}^MD^Z_i)$
   \STATE
   \STATE Construct pseudo-datasets for each learnt sub-domain:
   \FOR{$i \in [1, \ldots, M]$}
   \FOR{$j \in [1, \ldots, N_i]$}
   \STATE  $\hat{c}_{i,j} = \argmax_{c \in [1,\ldots,K]} [p(c | \bm{z}_{i,j}, \bm{m}_k^{\text{VAR}}, \kappa_k^{\text{VAR}}, \nu_k^{\text{VAR}}, \bm{W}_k^{\text{VAR}}, \bm{\alpha}^{\text{VAR}})]$
   \ENDFOR
   \ENDFOR
   \STATE 
   \FOR{$k \in [1, \ldots, K]$}
   \STATE $\mathcal{S}_k = \{\bm{x}_{i,j} | \hat{c}_{i,j} = k\}$
   \ENDFOR
   \STATE 
   \STATE Train per-sub-domain LoRA modules:
   \FOR{$k \in [1, \ldots, K]$}
   \STATE $\text{LoRA}_k = \text{TrainLoRA}(\mathcal{S}_k, \text{TSFM})$
   \ENDFOR
   \STATE
   \STATE Return per-sub-domain LoRA modules and Bayesian GMM:
   \STATE {\bfseries Return} $[\text{LoRA}_k]_{k=1}^{K}$, $[\bm{m}_k^{\text{VAR}}, \kappa_k^{\text{VAR}}, \nu_k^{\text{VAR}}, \bm{W}_k^{\text{VAR}}]_{k=1}^{K}, \; \bm{\alpha}^{\text{VAR}}$
   
\end{algorithmic}
\end{algorithm}




\clearpage

\FloatBarrier

\section{Additional Experimental Details}
\label{appen:ExpDetails}
While we have presented the most important experimental details in the main paper, there are some others needing mentioning. First, in our experiments, we set MixFTs Bayesian GMM hyper-parameters as, 
\begin{align}
    \bm{m} &= \text{Mean}(\bigcup_{i=1}^MD^Z_i)\\
    \kappa &= 1\\
    \nu &= \text{Dim}(Z)\\
    \bm{W} &= \text{Diag}(\text{Diag}(\text{Covariance}(\bigcup_{i=1}^MD^Z_i)))\\
    \bm{\alpha} &= \frac{1}{K}\bm{1},    
\end{align}
where $\text{Dim}(Z)$ is the dimensionality of the TSFMs embedding space and $\text{Diag}(\text{Diag}(.))$ takes a matrix and returns the matrix with only its diagonal entries, the off-diagonal entries being set to zero. The hyper-parameters were chosen as they are the common default values used by previous work. 
Second, we aim to keep all the hyper-parameters for the other fine-tuning methods to their default setup. Third, to ensure we tune LoRAs consistently across different methods we use the same LoRA hyper-parameters for all fine-tuning methods. The LoRA hyper-parameters used were hand tuned on the ETTh1 and ETTm1 datasets, performing well and were inspired by the ones used in \citep{gupta2024low}. More specifically, the LoRA hyper-parameters we use are $r=2$, $\alpha=16$, dropout $= 0.1$ and we also make use of OLoRA \citep{buyukakyuz2024olora} to improve low-rank training performance. To tune each LoRA module we use AdamW \citep{loshchilov2017decoupled}, a learning rate of $lr=0.00005$ and a batch size of $256$. Forth, to convert the tokens given by the TSFMs to the representations used for the Bayesian GMM, we construct a vector formed of the mean of each token. Last, given we are interested in zero-shot forecasting, we do not scale the datasets by using a training split like some previous work \citep{nietime2022}. Instead, we rely on the TSFMs internal scaling mechanism. For both Chronos Bolt and Moirai-1.1-R this is reversible instance norm (RevIN) \citep{kim2021reversible}.

In the main text, we mention to improve LoRA fine-tuning on time series we make use of \textbf{a)} MixUp \citep{zhang2017MixUp} and \textbf{b)} replaying pretraining data. Here, we describe our use of them in more detail. For MixUp, we use the standard dynamic variant, mixing each batch as given in \citet{zhang2017MixUp}. However, to ensure that time series of different scales do not skew the mixing we scale the data, as in \citet{Ansari2024Chronos}, using RevIN. For replaying the pretraining data, for each training batch of fine-tuning data, we sample the same number of data points from the pretraining data as the batch. This replay data is then append to the training batch. We append the replay data before we use MixUp, so the pretraining data is also mixed with the fine-tuning data. We use Chronos Bolt's pretraining data for both TSFMs looked at, as we could not fully identify what data Moirai-1.1-R was pretrained on.


\clearpage

\section{Additional Experimental Results}

\subsection{Selection of Number of Mixture Components}
\label{appen:K_selection}
\begin{table*}[h]
\centering
\caption{\textbf{MixFT validation performance for different number of components $K$.} The table contains the average rank of MixFT using different number of components. The average is taken over the validation sets of the fine-tuning datasets. The table shows that $K=2$ has the lowest average rank and hence should be selected.}
\label{table:K_selection}
\begin{tabular}{@{}c@{\hskip 1mm}|@{\hskip 1mm}c@{}}
\toprule
\addlinespace
\multirow{2}{*}{Num. of Components} & \multirow{2}{*}{Val. Avg. Rank ($\mathbf{\downarrow}$)} \\
\\
\midrule
$K = 2$ & \textbf{2.17} \\
$K = 3$ & 3.17 \\
$K = 4$ & \underline{2.83} \\
$K = 5$ & 3.33 \\
$K = 10$ & 3.50 \\
\bottomrule
\end{tabular}
\end{table*}
To select the number of mixture components used by MixFT we can use the validation performance on the fine-tuning data, as is standard practice \citep{ostapenko2024towards}. Specifically, we use the average rank as our performance metric and use the last $10\%$ of each fine-tuning dataset as validation data. We present the validation performance results in Table~\ref{table:K_selection}. The table show that using two components ($K=2$) gives the best validation performance. This result justifies our use of $K=2$ in our main experiments. Additionally, it is interesting to contrast the validation performances of MixFT using different number of components with their respective performances on the evaluation datasets, which is shown in Table~\ref{table:k}. We find that there is general agreement between the validation and evaluation performance, with $K=2$ achieving the best performance for both. This suggests our method of selecting the number of mixture components leads to good performance on the evaluation datasets. However, as expected, validation performance does not fully predict the relative evaluation performance of using each number of components. For instance, the ordering of $K=3$ and $K=4$ in terms of the validation and evaluation performance is different.

\FloatBarrier
\subsection{Ablation of Zero-Shot Mixture Selection}
\label{appen:selection}
\begin{table*}[h]
\centering
\caption{\textbf{MASE of versions of MixFT with differing component selection methods, when zero-shot forecasting using Chronos Bolt.}  The table shows MixFT with $\mu$ and Arrow routing which are used by some of the other fine-tuning methods compared to in the main paper. Also the table shows two methods which use the probabilities given by MixFTs Bayesian GMM to perform component selection, soft and ensemble-MixFT. In the table we bold and underline the best and second best results, respectively. We find that that MixFT performs the best or second best for all time series, bar CloudD1, and achieves the lowest average rank. This shows that MixFTs way of selecting what LoRA to use for zero-shot forecasting performs well.}
\label{table:routing}
\begin{tabular}{@{}c@{\hskip 1mm}|@{\hskip 1mm}ccccc@{}}
\toprule
\addlinespace
\multicolumn{1}{c}{} & \multicolumn{5}{c}{MixFT with Different Component Selection Methods} \\
\cmidrule(l){2-6}
\multirow{2}{*}{Datasets} & \multirow{2}{*}{\begin{tabular}{c} $\mu$ \\ MixFT \\ \end{tabular}} &  \multirow{2}{*}{\begin{tabular}{c} Arrow \\ MixFT \\ \end{tabular}} &  \multirow{2}{*}{\begin{tabular}{c} Soft \\ MixFT \\ \end{tabular}} & \multirow{2}{*}{\begin{tabular}{c} Ensemble \\ MixFT \\ \end{tabular}}  &  \multirow{2}{*}{MixFT (ours)} \\ 
\\
\midrule
CloudD1  & $\mathbf{1.411_{\pm{0.003}}}$ & $1.594_{\pm{0.055}}$ & $1.697_{\pm{0.009}}$ & \underline{$1.480_{\pm{0.003}}$} & $1.485_{\pm{0.008}}$ \\
CloudD2  & $1.425_{\pm{0.001}}$ & $1.384_{\pm{0.006}}$ & $1.401_{\pm{0.010}}$ & \underline{$1.351_{\pm{0.001}}$} & $\mathbf{1.343_{\pm{0.004}}}$ \\
\addlinespace
BizITObs-L2C  & $\underline{3.087_{\pm{0.002}}}$ & $3.163_{\pm{0.013}}$ & $3.237_{\pm{0.012}}$ & $3.088_{\pm{0.003}}$ &  $\mathbf{3.083_{\pm{0.006}}}$ \\
BizITObs-App  & $1.907_{\pm{0.016}}$ & $1.277_{\pm{0.038}}$ & $1.168_{\pm{0.014}}$ & \underline{$1.015_{\pm{0.002}}$} &  $\mathbf{0.989_{\pm{0.024}}}$ \\
US-Births & $0.973_{\pm{0.001}}$ & $0.977_{\pm{0.002}}$ & $0.968_{\pm{0.001}}$ & \underline{$0.946_{\pm{0.001}}$} &  $\mathbf{0.942_{\pm{0.002}}}$ \\
\addlinespace
M4-Daily & $\mathbf{7.172_{\pm{0.001}}}$ & $7.450_{\pm{0.024}}$ & $7.535_{\pm{0.032}}$ & \underline{$7.238_{\pm{0.005}}$} &  $\underline{7.238_{\pm{0.005}}}$ \\
M4-Monthly & $\underline{7.639_{\pm{0.007}}}$ & $7.780_{\pm{0.008}}$ & $7.911_{\pm{0.052}}$  & $7.690_{\pm{0.043}}$ &  $\mathbf{7.638_{\pm{0.014}}}$\\
M4-Quarterly & $\mathbf{8.100_{\pm{0.008}}}$ & $8.344_{\pm{0.017}}$ & $8.423_{\pm{0.020}}$ & $8.140_{\pm{0.024}}$ &  $\underline{8.135_{\pm{0.010}}}$ \\
\addlinespace
ETTh2 & $\mathbf{1.381_{\pm{0.001}}}$ & $1.411_{\pm{0.002}}$ & $1.413_{\pm{0.001}}$ & $1.389_{\pm{0.001}}$ &  $\underline{1.387_{\pm{0.001}}}$ \\
ETTm2 & $\mathbf{0.723_{\pm{0.001}}}$ & $0.749_{\pm{0.001}}$ & $0.747_{\pm{0.001}}$ & \underline{$0.730_{\pm{0.001}}$} &  $\underline{0.730_{\pm{0.001}}}$ \\
\addlinespace
\midrule
\rowcolor[gray]{0.8}  \textbf{Avg. Rank ($\mathbf{\downarrow}$)} & \underline{$2.3$} & $4.1$ & $4.4$ & $2.4$ & $\bm{1.6}$ \\
\bottomrule
\end{tabular}
\end{table*}
To analyse the benefit of using MixFTs zero-shot mixture selection method, we compare it to other selection methods. First, we compare it to when using the mixture selection methods of the other fine-tuning methods looked at in this work, $\mu$ and Arrow  routing. By doing this, we obtain the alternative MixFT versions $\mu$-MixFT and Arrow-MixFT. Second, we compare MixFTs \textit{hard} selection mechanism to soft ones using the probabilities given by the Bayesian GMM, instead of argmaxing over them. Soft-MixFT is when we take a weighted average of the per-sub-domain LoRA modules. We use the probabilities of the given context belonging to a sub-domain as that sub-domains LoRAs weight. Ensemble-MixFT is when, instead of averaging the LoRA modules, we average the forecasts given by using each per-sub-domain LoRA module with the TSFM. Again, the weights used are the probabilities the given context belongs to each sub-domain. 

The results of this ablation are shown in Table~\ref{table:routing}, showing among other things that MixFT selection mechanism performs the best. The results show that Arrow and Soft selection mechanisms perform the worse, obtaining similar average ranks. While $\mu$ and Ensemble MixFT perform better. This is interesting as $\mu$ routing is more simplistic than Arrow routing but performs better than it for MixFT, while it performs worse than it for per-dataset methods. This suggests that Arrow routing is somehow well fit to per-dataset methods, while it struggles with MixFT. The results also show that MixFTs hard selection mechanism performs the best. This is because, it has the lowest average rank and that it performs best or second best for all datasets bar CloudD1. This justifies the use of MixFTs zero-shot selection mechanism. Additionally, the results suggests, that as soft methods do not perform as well as MixFTs hard one, converting probabilities to weighting of LoRA modules may not be straight forward. We think this is an interesting direction for future work. 



\subsection{Ablation of Mixture Model}
\label{appen:mixmodel}
\begin{table*}[h]
\centering
\caption{\textbf{MASE of versions of MixFT with differing probabilistic models, when using Chronos Bolt.} MixFT uses a Bayesian GMM to identify and learn sub-domains. In the table we show its performance using a range of different probabilistic models, to identify and learn sub-domains. In the table we bold and underline the best and second best results, respectively. We find that using a Bayesian GMM performs best or second best for all datasets, bar M4-Daily, and achieves the lowest average rank. This shows it is a good choice for the probabilistic model of MixFT. }
\label{table:model}
\begin{tabular}{@{}c@{\hskip 1mm}|@{\hskip 1mm}ccc@{}}
\toprule
\addlinespace
\multicolumn{1}{c}{}  & \multicolumn{3}{c}{MixFT with Different Prob. Models} \\
\cmidrule(l){2-4}
\multirow{2}{*}{Datasets} & \multirow{2}{*}{K-means}  &  \multirow{2}{*}{Topic Model} &  \multirow{2}{*}{Bayesian GMM (ours)} \\ 
\\
\midrule
CloudD1 &  $\underline{1.516_{\pm{0.005}}}$ & $1.532_{\pm{0.004}}$ & $\mathbf{1.485_{\pm{0.008}}}$ \\
CloudD2 &  $1.363_{\pm{0.003}}$ & $\mathbf{1.328_{\pm{0.009}}}$ & $\underline{1.343_{\pm{0.004}}}$ \\
\addlinespace
BizITObs-L2C &  $\underline{3.096_{\pm{0.005}}}$ & $3.100_{\pm{0.002}}$ &   $\mathbf{3.083_{\pm{0.006}}}$ \\
BizITObs-App  &  $1.048_{\pm{0.006}}$ & $\underline{1.030_{\pm{0.041}}}$ &  $\mathbf{0.989_{\pm{0.024}}}$ \\
US-Births & $0.967_{\pm{0.005}}$ & $\underline{0.960_{\pm{0.009}}}$ &   $\mathbf{0.942_{\pm{0.002}}}$ \\
\addlinespace
M4-Daily & $\mathbf{7.230_{\pm{0.004}}}$ & $\underline{7.237_{\pm{0.008}}}$ &   $7.238_{\pm{0.005}}$ \\
M4-Monthly &  $7.737_{\pm{0.006}}$ & $\mathbf{7.552_{\pm{0.002}}}$ &  $\underline{7.638_{\pm{0.014}}}$\\
M4-Quarterly &  $\mathbf{8.131_{\pm{0.003}}}$ & $8.238_{\pm{0.014}}$ &   $\underline{8.135_{\pm{0.010}}}$ \\
\addlinespace
ETTh2 &  $1.390_{\pm{0.001}}$ & $\mathbf{1.387_{\pm{0.001}}}$ &  $\mathbf{1.387_{\pm{0.001}}}$ \\
ETTm2 &  $\underline{0.730_{\pm{0.001}}}$ & $\mathbf{0.728_{\pm{0.001}}}$ &  $\underline{0.730_{\pm{0.001}}}$ \\
\addlinespace
\midrule
\rowcolor[gray]{0.8}  \textbf{Avg. Rank ($\mathbf{\downarrow}$)} & $2.3$ & \underline{$1.9$} & $\bm{1.6}$  \\
\bottomrule
\end{tabular}
\end{table*}
To explore the benefits of using a Bayesian GMM to model sub-domains instead of another probabilistic model, we perform an ablation. We compare MixFT use of a Bayesian GMM with alternate versions of MixFT: one which uses a GMM fit with K-means and another which uses a topic model \citep{vayansky2020review}. The specific topic model we use is an adjusted version of the one proposed by \citep{dieng2020topic}. The changes we make are \textbf{a)} instead of learning an embedding we use the TSFMs embedding and \textbf{b)} given our "words" (i.e. contexts) are continuous we use Gaussians as their distribution, given a document and topic assignment. 

The results of this ablation are presented in Table~\ref{table:model}, showing the Bayesian GMM performs best. More specifically, in terms of average rank, we find that using K-means performs worse. This is probably due to overfitting to some shallow local minima as it is known to do \citep[Chapter~9]{BishopPRML2006}. While using the Topic model performs second best, performing somewhat similarly to the Bayesian GMM. This is not surprising, as the topic model is very similar to the Bayesian GMM but for the fact it has an added dependency over datasets. It is interesting however, that this added dependency, regulating what sub-domains appear in each dataset, does not help. We see it as future work to see how this can be improved, leveraging dataset labels somewhat, while not becoming a full per-dataset method. Finally, we find that the Bayesian GMM performs the best. This validates our use of it as MixFTs probabilistic model for the sub-domains.

\FloatBarrier


\subsection{Ablation on the Number of Mixture Components}
\label{appen:components}
\begin{table*}[h!]
\centering
\caption{\textbf{MASE of using MixFT with differing number of mixture components, when using Chronos Bolt.} In the table we show when using MixFT with one to ten mixture components. Using one is equivalent to  Shared and using two is what MixFT uses in the rest of our experiments. In the table we bold and underline the best and second best results, respectively. We find that different numbers of mixture components often perform similarly and that using two is best in terms of average rank.}
\label{table:k}
\begin{tabular}{@{}c@{\hskip 1mm}|@{\hskip 1mm}cccccc@{}}
\toprule
\addlinespace
\multicolumn{1}{c}{}  & \multicolumn{6}{c}{MixFT with Different Numbers of Components } \\
\cmidrule(l){2-7}
\multirow{2}{*}{Datasets} & \multirow{2}{*}{$K=1$ (Shared)}  &  \multirow{2}{*}{$K=2$ (ours)} &  \multirow{2}{*}{$K=3$} &  \multirow{2}{*}{$K=4$} & \multirow{2}{*}{$K=5$} &  \multirow{2}{*}{$K=10$} \\ 
\\
\midrule
CloudD1  & $1.553_{\pm{0.024}}$ & $\underline{1.485_{\pm{0.008}}}$ & $1.500_{\pm{0.010}}$ & $1.489_{\pm{0.002}}$ & $1.518_{\pm{0.017}}$ & $\mathbf{1.448_{\pm{0.004}}}$ \\
CloudD2  & $\mathbf{1.342_{\pm{0.004}}}$ & $\underline{1.343_{\pm{0.004}}}$ & $1.375_{\pm{0.010}}$ & $1.365_{\pm{0.003}}$ & $1.384_{\pm{0.007}}$ & $1.423_{\pm{0.003}}$ \\
\addlinespace
BizITObs-L2C  & $3.101_{\pm{0.011}}$ & $3.083_{\pm{0.006}}$ & $\underline{3.061_{\pm{0.009}}}$ & $\mathbf{3.058_{\pm{0.009}}}$ & $3.068_{\pm{0.007}}$ & $3.065_{\pm{0.010}}$ \\
BizITObs-App & $1.060_{\pm{0.020}}$ & $0.989_{\pm{0.024}}$ & $1.001_{\pm{0.018}}$ & $\mathbf{0.970_{\pm{0.013}}}$ & $\underline{0.985_{\pm{0.019}}}$ & $1.127_{\pm{0.062}}$ \\
US-Births  & $\underline{0.967_{\pm{0.007}}}$ & $\mathbf{0.942_{\pm{0.002}}}$ & $1.007_{\pm{0.005}}$ & $1.034_{\pm{0.006}}$ & $1.025_{\pm{0.001}}$ & $1.012_{\pm{0.002}}$ \\
\addlinespace
M4-Daily  & $7.254_{\pm{0.016}}$ & $7.238_{\pm{0.005}}$ & $\mathbf{7.233_{\pm{0.004}}}$ & $\underline{7.236_{\pm{0.003}}}$ & $7.244_{\pm{0.002}}$ & $7.245_{\pm{0.007}}$ \\
M4-Monthly  & $\mathbf{7.580_{\pm{0.042}}}$ & $\underline{7.638_{\pm{0.014}}}$ & $7.707_{\pm{0.041}}$ & $7.739_{\pm{0.008}}$ & $7.760_{\pm{0.012}}$ & $7.708_{\pm{0.017}}$ \\
M4-Quarterly & $8.269_{\pm{0.037}}$ & $\mathbf{8.135_{\pm{0.010}}}$ & $\underline{8.154_{\pm{0.015}}}$ & $8.188_{\pm{0.006}}$ & $8.183_{\pm{0.024}}$ & $8.169_{\pm{0.021}}$ \\
\addlinespace
ETTh2  & $1.388_{\pm{0.001}}$ & $1.387_{\pm{0.001}}$ & $1.391_{\pm{0.001}}$ & $1.389_{\pm{0.001}}$ & $\underline{1.385_{\pm{0.001}}}$ & $\mathbf{1.383_{\pm{0.001}}}$ \\
ETTm2  & $0.731_{\pm{0.003}}$ & $0.730_{\pm{0.001}}$ & $0.730_{\pm{0.001}}$ & $0.729_{\pm{0.001}}$ & $\underline{0.728_{\pm{0.001}}}$ & $\mathbf{0.727_{\pm{0.001}}}$ \\
\addlinespace
\midrule
\rowcolor[gray]{0.8}  \textbf{Avg. Rank ($\mathbf{\downarrow}$)} & $4.3$ & $\bm{2.6}$ &\underline{$3.3$} & $3.4$ & $3.9$ & $3.4$ \\
\bottomrule
\end{tabular}
\end{table*}

\FloatBarrier
\clearpage

\subsection{Analysis of Identified Sub-Domains}
\subsubsection{How Confident is MixFT's Sub-Domain Classification}
\label{appen:ent}
\begin{table*}[h]
\setlength\tabcolsep{2.2pt}
\centering
\caption{\textbf{Confidence of MixFTs classification of a contexts sub-domain in terms of average classification entropy.} The average classification entropy of a dataset $D$ is $\mathbb{E}_{\bm{x} \sim D}[H(c|\bm{z} = \text{TSFM}_{embed}(\bm{x}))]$ and measures the average uncertainty in the classification of $c$ for $\bm{x} \in D$. We find that the average entropy is small across all fine-tuning and evaluation datasets, where an entropy of $0.01$ bit corresponds to a $0.999$ estimated probability a context is of a particular sub-domain. This shows that MixFT is on average very confident what contexts belong to what sub-domains. }
\label{table:results_ent}
\begin{tabular}{@{}cc@{\hskip 1mm}|@{\hskip 1mm}c@{}}
\toprule
\addlinespace
\multirow{2}{*}{} & \multirow{2}{*}{Datasets} &  \multirow{2}{*}{\begin{tabular}{c} Average Classification Entropy \\ $\mathbb{E}_{\bm{x} \sim D}[H(c|\bm{z} = \text{TSFM}_{embed}(\bm{x}))]$ \end{tabular}} \\ 
\multirow{2}{*}{} & \\
\midrule
\multirow{6}{*}{\STAB{\rotatebox[origin=c]{90}{\textbf{Fine-tuning}}}} 
& CloudD3 &  0.0095 \\
& CloudD4 & 0.0075 \\
\addlinespace
& BizITObs-Service &  0.0053\\
& BitBrains Fast Storage &  0.0101\\
\addlinespace
& M4-Hourly &  0.0093 \\
& M4-Weekly & 0.0050 \\
\addlinespace
\midrule
\rowcolor[gray]{0.8}  & \textbf{Average} &  \textbf{0.0078} \\

\midrule
\addlinespace

\multirow{12}{*}{\STAB{\rotatebox[origin=c]{90}{\textbf{Evaluation}}}} 
& CloudD1 &  0.0124 \\
& CloudD2 &  0.0110 \\
\addlinespace
& BizITObs-L2C & 0.0054 \\
& BizITObs-App & 0.0037 \\
& US-Births & 0.0131 \\
\addlinespace
& M4-Daily &  0.0062 \\
& M4-Monthly &  0.0054 \\
& M4-Quarterly & 0.0068 \\
\addlinespace
& ETTh2 &  0.0168 \\
& ETTm2 &  0.0145 \\
\addlinespace
\midrule
\rowcolor[gray]{0.8} & \textbf{Average} & \textbf{0.0095} \\
\bottomrule
\end{tabular}
\end{table*}
To see how confident MixFT is at classifying a contexts sub-domain, in Table~\ref{table:results_ent} we present the average classification entropy $\mathbb{E}_{\bm{x} \sim D}[H(c|\bm{z} = \text{TSFM}_{embed}(\bm{x}))]$ for each dataset $D$ used in our experiments. Given that we use two components for MixFT, the maximum value of $H(c|\bm{z} = \text{TSFM}_{embed}(\bm{x}))$ for some $\bm{x}$ is $1$ bit and a value of less than $0.01$ means that $p(c | \bm{z},  \phi_c^{\text{VAR}}, \bm{\alpha}^{\text{VAR}}) \geq 0.999$ for the most likely component $c$. Therefore, as the average entropy for each dataset displayed in Table~\ref{table:results_ent} is near or below $0.01$, it shows that, on average, MixFT is very confident in classifying the sub-domain of a context. This finding holds for both the fine-tuning datasets used to fit the Bayesian GMM and the evaluation datasets, with the datasets of the latter having slightly larger average entropy in general.  

One reason for looking at the confidence of MixFT in classifying a context's sub-domain is to explore whether using only the most likely sub-domain's LoRA module for forecasting is sensible. This is because, if there is multiple likely candidates for a context's sub-domain then using only the most likely sub-domain's LoRA module is less sensible than using a mixture of LoRAs. However, given that this is not the case, as usually only one sub-domain is likely for a given context (see Table~\ref{table:results_ent}), using only the most likely sub-domains LoRA module in forecasting makes sense. Additionally, as shown in the next section, it also makes MixFT robust to inaccurate forecasts given when using LoRA modules of unlikely sub-domains.         

\clearpage

\subsubsection{Characteristics of Identified Sub-Domain Forecasts}
\label{appen:topic_forecasts}
\begin{figure}[h]
    \centering
    \begin{minipage}{0.49\textwidth}
        \centering
        \includegraphics[width=\textwidth]{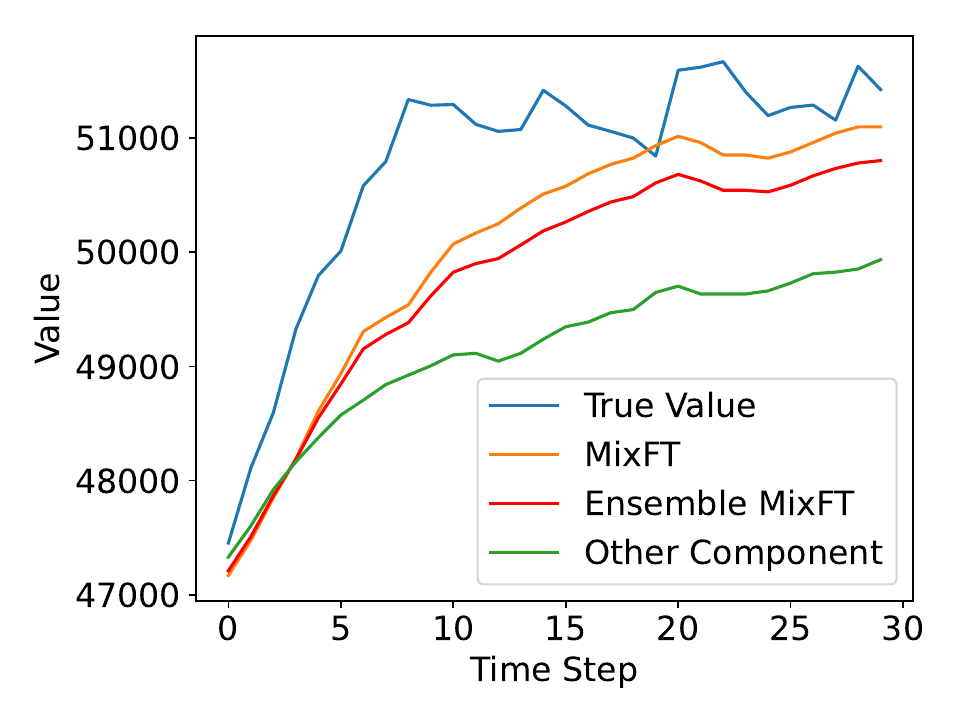}
        \caption*{\textbf{a) }}
     \end{minipage}\hfill
    \begin{minipage}{0.49\textwidth}
        \centering
        \includegraphics[width=\textwidth]{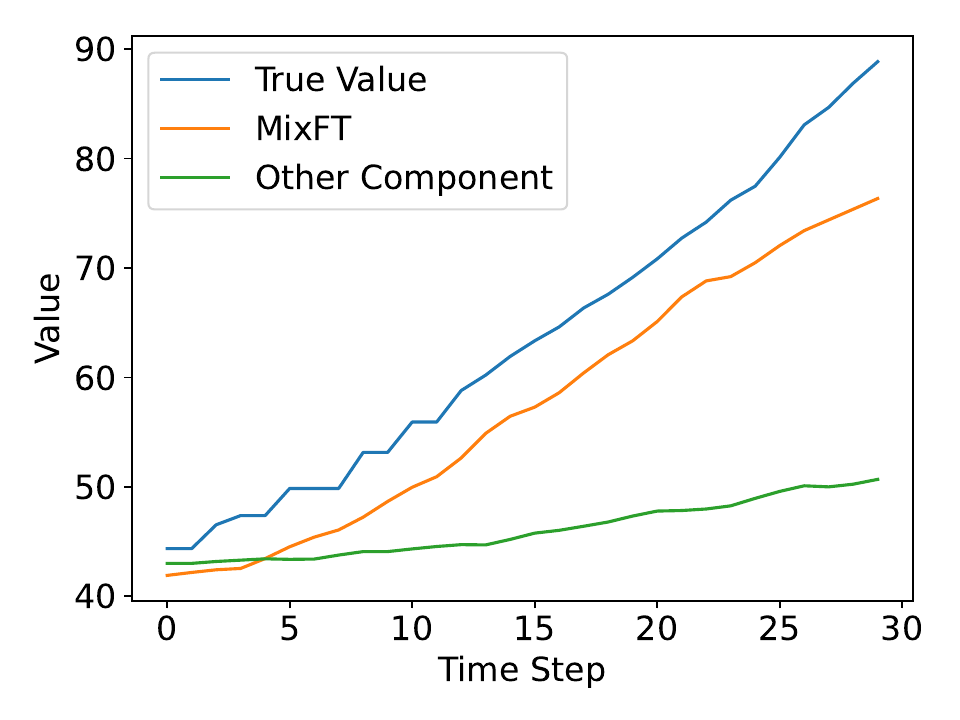}
        \caption*{\textbf{b) }}
    \end{minipage}
    \caption{\textbf{Example forecasts on BizITObs-Application given by MixFT and when using the other sub-domains LoRA module in a mixture or on its own.} The two plot shows forecasts for MixFT, Ensemble MixFT (see Appendix~\ref{appen:selection}) and the forecast given by using the LoRA module not selected by MixFT, calling this \textit{Other Component} in the plots. The plot also present the true value. In plot \textbf{b)} we do not show the forecasts of Ensemble MixFT as it is identical to that of MixFT---MixFT classifies with probability $1.0$ that the given context belongs to one of the sub-domains. The plots show that using the LoRA module MixFT selects gives better forecasts than using the one it does not. Furthermore, that ensembleling both forecasts can result in worse performance than MixFT because the \textit{Other Component} forecast can be inaccurate given that it is forecasting an OOD context for its sub-domain.}
    \label{fig:topic_forecasts}
\end{figure}
To examine the different forecasts given by using the different LoRA modules learnt by MixFT, in Figure~\ref{fig:topic_forecasts} we plot two examples. We display the forecasts of \textit{MixFT}; \textit{Ensemble MixFT} which is when we take a weighted average of the forecasts given by the LoRA modules of each sub-domain, weighted by the probability the context is from each sub-domain; \textit{Other Component} which is the forecast given by using the LoRA module of the sub-domain not chosen by MixFT; and the \textit{true value}. The figure shows that MixFT performs better than Ensemble MixFT, which as shown in Table~\ref{table:routing} holds in general as well, evidencing the reason of using hard sub-domain assignment in MixFT. A reason for the superior performance of MixFT is shown in Figure~\ref{fig:topic_forecasts}, as the forecasts made when using the LoRA module of the sub-domain not selected by MixFT are inaccurate. This is likely due to the OOD nature of the context for that sub-domain, as that sub-domains LoRA module has not been trained on similar contexts and hence gives a bad forecast. Therefore, mixing this bad forecast with the one given when using the LoRA module of the in-distribution sub-domain reduces performance. This demonstrates a reason why using hard sub-domain assignment performs better than soft ones for MixFT. 

\FloatBarrier

\subsubsection{Characteristics of Identified Sub-Domain Contexts}
\label{Appen:contexts}
\begin{figure}[h]
    \centering
    \begin{minipage}{0.49\textwidth}
        \centering
        \includegraphics[width=\textwidth]{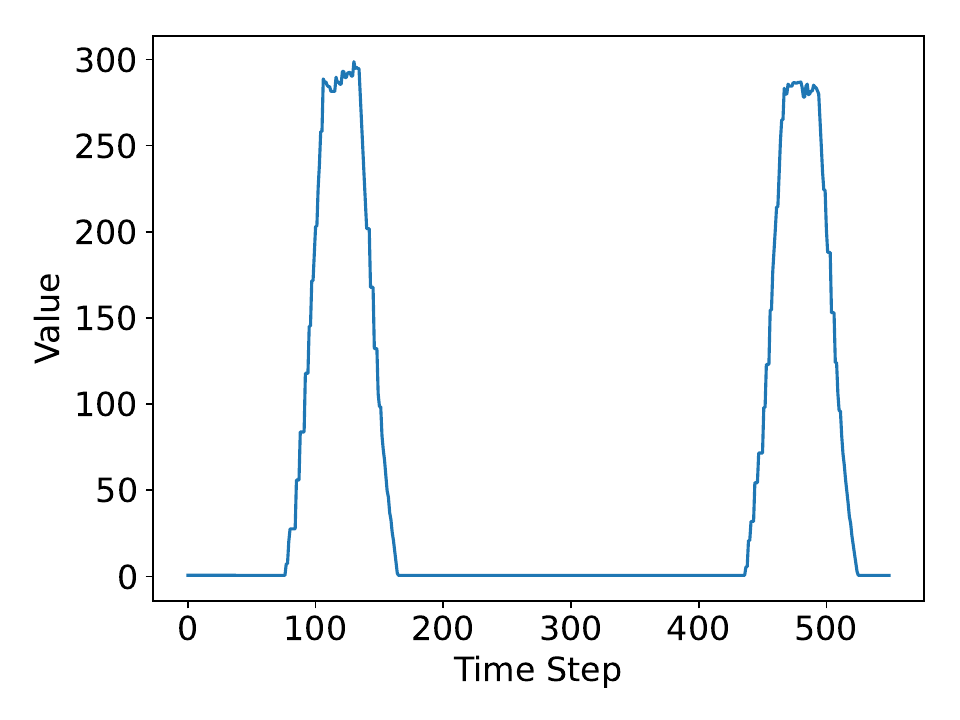}
        \caption*{\textbf{a) A context identified as being from the first sub-domain}}
     \end{minipage}\hfill
    \begin{minipage}{0.49\textwidth}
        \centering
        \includegraphics[width=\textwidth]{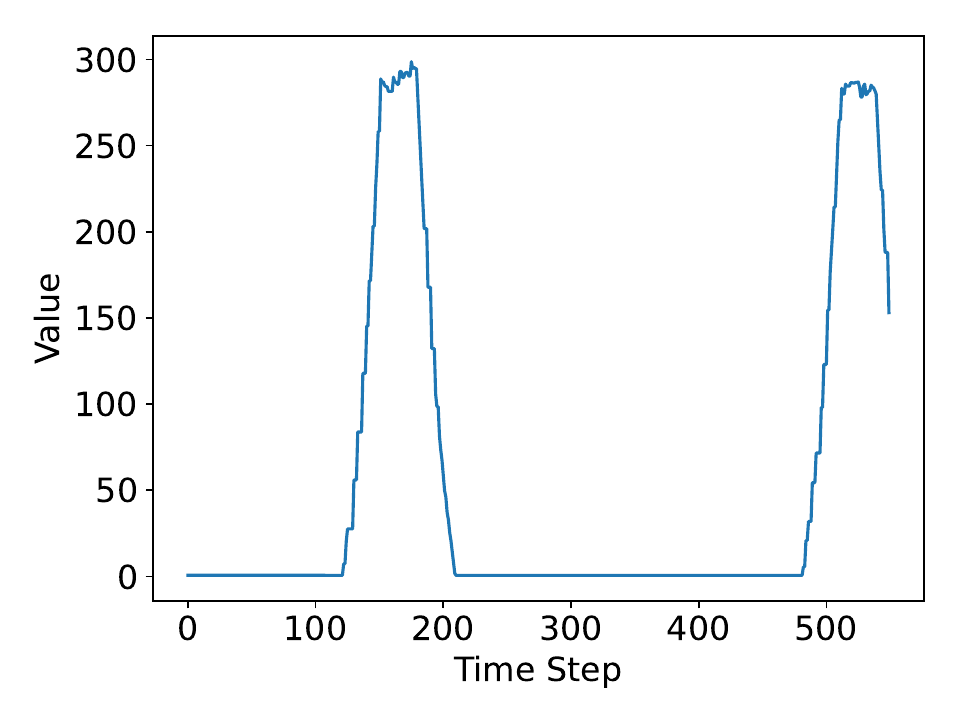}
        \caption*{\textbf{b) A context identified as being from the second sub-domain}}
    \end{minipage}
    \caption{\textbf{Plots of two contexts identified as belonging to the two mixture components of MixFT, respectively, for the first channel of the fine-tuning dataset BizITObs-service, when using Chronos Bolt.} The plot shows one of the characteristics of the learnt sub-domains, in that when the end of a context is varying (i.e. is spiking in the figure) then it is seen as belonging to the second component, else it belongs to the first. This is the reason why in Figure~\ref{fig:mixDist}, for BizITObs-service there is a periodic pattern as it has periodic spikes. This figure demonstrates that MixFTs found sub-domains are interpretable and understandable.}
    \label{fig:topic_contexts1}
\end{figure}

\begin{figure}[h]
    \centering
    \begin{minipage}{0.49\textwidth}
        \centering
        \includegraphics[width=\textwidth]{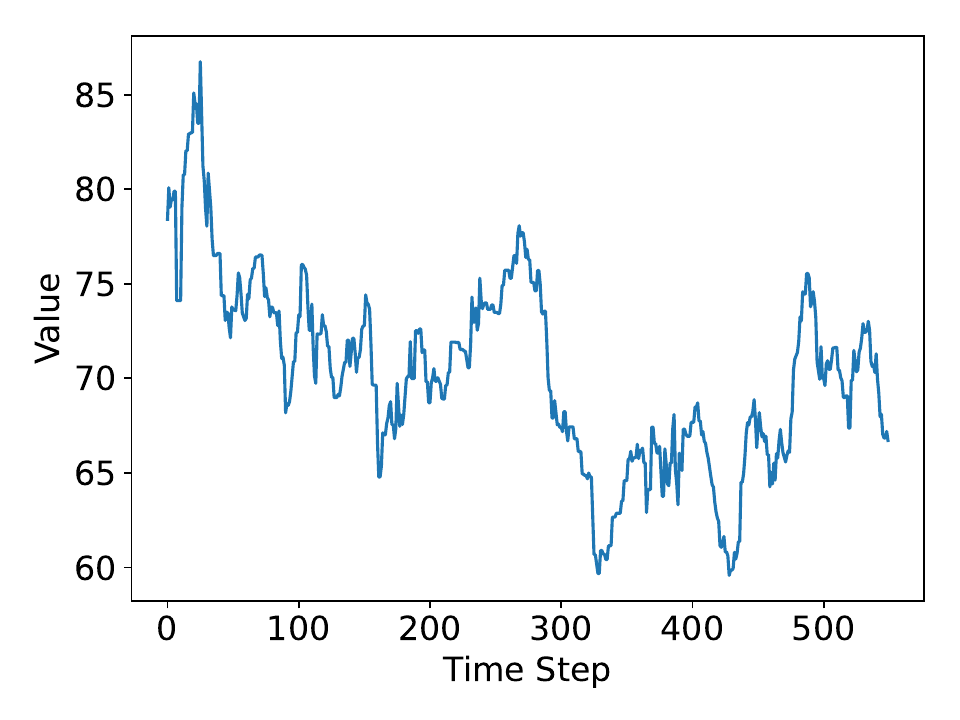}
        \caption*{\textbf{a) A context identified as being from the first sub-domain}}
     \end{minipage}\hfill
    \begin{minipage}{0.49\textwidth}
        \centering
        \includegraphics[width=\textwidth]{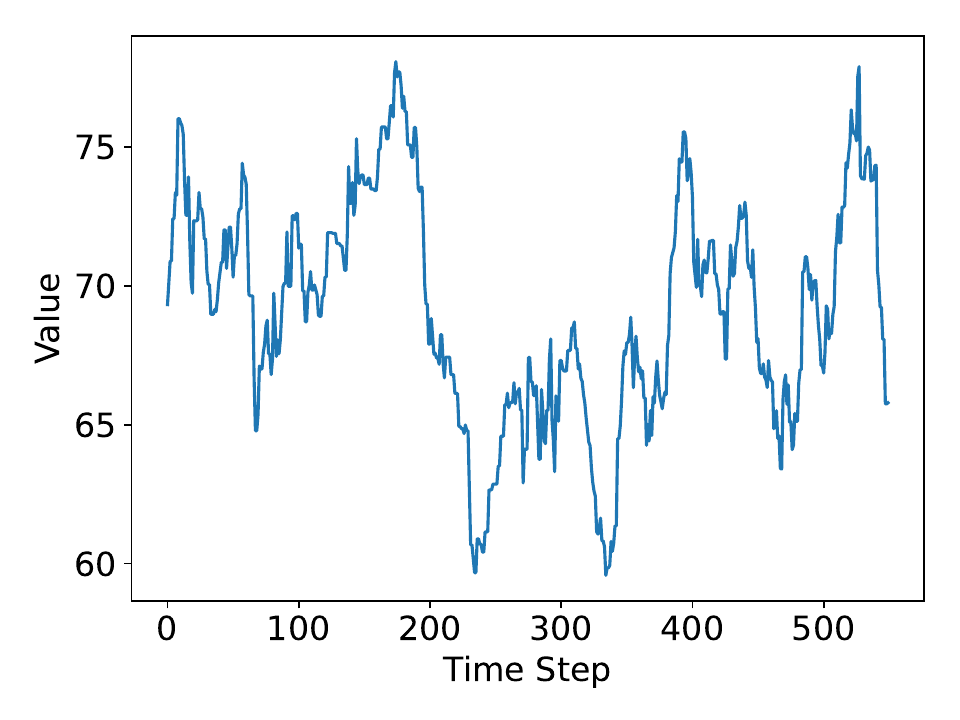}
        \caption*{\textbf{b) A context identified as being from the second sub-domain}}
    \end{minipage}
    \caption{\textbf{Plots of two contexts identified as belonging to the two mixture components of MixFT, respectively, for the eighth channel of the fine-tuning dataset BizITObs-service, when using Chronos Bolt.} Unlike Figure~\ref{fig:topic_contexts1}, these contexts do not have a clear seasonal pattern. Because of this, it is harder to identify the features used to decide the sub-domain. However, it is possible trend is used, as plot \textbf{a)} has a negative trend while plot \textbf{b)} has no discernable trend.  }
    \label{fig:topic_contexts3}
\end{figure}

To understand what MixFTs mixture model has identified as a sub-domain, we have plotted examples of contexts belonging to each sub domain. The examples are shown in Figure~\ref{fig:topic_contexts1} and \ref{fig:topic_contexts3}. In Figure~\ref{fig:topic_contexts1}, we find that for the first channel of BizITObs-service that contexts belonging to the first mixture component do not vary at the end of the context (i.e. are a flat line), while for the second mixture component they do. This is the reason why in Figure~\ref{fig:mixDist} for BizITObs-service there is a periodic pattern. Therefore, the discovered sub-domains might be partially selected based on the near pasts signal volatility. This, among other things, sheds light on how Chronos bolt choses to embed contexts for the first channel of BizITObs-service, where it seems to group contexts of similar near pasts volatility together. In Figure~\ref{fig:topic_contexts3}, we see that not all contexts have seasonal patterns like Figure~\ref{fig:topic_contexts1}, in these circumstances it is harder to understand what each sub-domain is identifying. However, we see that one of the contexts in the figure has a downwards trend while the other does not, this might be a feature use to classify sub domains.   

\FloatBarrier
\clearpage

\subsubsection{Selection of LoRA Modules for MixFT when Zero-Shot Forecasting}
\begin{figure}[h!]
    \centering
    \includegraphics[scale=0.6]{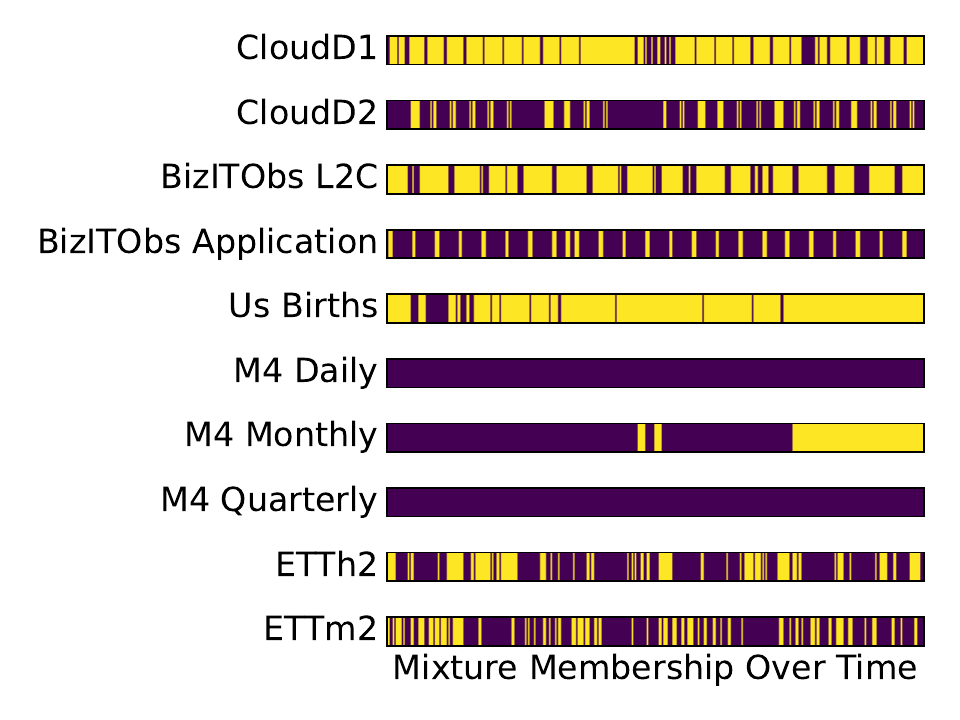}
    \caption{\textbf{Mixture membership of the evaluation datasets for MixFT, when using Chronos Bolt.} The plot shows, for the first channel of each evaluation dataset, when the time series is learnt to be of the first mixture component (purple area) or the second component (yellow area). We find, as expected, that there are less consistent patterns of mixture membership than in the fine-tuning data (see Figure~\ref{fig:mixDist}). This demonstrates the difficultly of identifying sub-domains from zero-shot data. However, there are still patterns of usage. For example, for BizITObs-Application we see a periodic pattern. This suggests while it is hard to identify zero-shot mixture membership, MixFT still does a reasonable job and this is one of the reasons it performs well in our experiments.}
    \label{fig:testMixDist}
\end{figure}


    



\end{document}